\documentclass{article}
\usepackage{PRIMEarxiv}

\usepackage[utf8]{inputenc} 
\usepackage[T1]{fontenc}    
\usepackage{hyperref}       
\usepackage{url}            
\usepackage{booktabs}       
\usepackage{amsfonts}       
\usepackage{nicefrac}       
\usepackage{microtype}      
\usepackage{lipsum}
\usepackage{natbib} 
\usepackage{fancyhdr}       
\usepackage{graphicx}       
\graphicspath{{media/}}     
\usepackage{amsmath}
\usepackage{pdfpages}
\usepackage{caption}
\usepackage{titlesec}

\usepackage{tcolorbox}
\tcbuselibrary{skins, breakable}

\usepackage{ragged2e}
\usepackage{ragged2e}
\usepackage{microtype}
\usepackage{orcidlink}
\usepackage{tikz}
\usetikzlibrary{positioning,arrows.meta,shapes.geometric,backgrounds,fit,calc}

\usepackage{mdframed}

\newtcolorbox{Box1}[2][]{
  lower separated=true,
  colback=black!5!white,
  colframe=black,
  fonttitle=\bfseries,
  colbacktitle=white!20!white,
  coltitle=black,
  enhanced,
  attach boxed title to top left={xshift=0.5cm, yshift=-2mm},
  title=#2,
  boxrule=0.9pt,
  boxed title style={colframe=black, boxrule=0.5pt},
  left=2mm,
  right=2mm,
  top=3mm,
  bottom=2mm,
  before upper={
    \footnotesize
    \ttfamily
    \justifying
    \setlength{\parindent}{0pt}
    \setlength{\emergencystretch}{3em}
  },
  #1}

\newtcolorbox{Box2}[2][]{
  lower separated=true,
  colback=white,
  colframe=black,
  fonttitle=\bfseries,
  colbacktitle=white,
  coltitle=black,
  enhanced,
  attach boxed title to top center={yshift=-2mm},
  title=#2,
  boxrule=1pt,
  boxed title style={colframe=black, boxrule=0.5pt},
  left=1mm,
  right=1mm,
  top=2mm,
  bottom=1mm,
  #1}

\title{Graph-Native Reinforcement Learning Enables Traceable Scientific Hypothesis Generation through Conceptual Recombination}

\author{
  \orcidlink{0000-0002-0532-9791} \textbf{Subhadeep Pal} \\
  Department of Civil and Environmental Engineering \\
  Massachusetts Institute of Technology \\
  Cambridge, MA, USA \\
  \And
  \orcidlink{0000-0002-0169-4003} \textbf{Shashwat Sourav} \\
  Department of Physics \\
  Washington University in St. Louis \\
  St. Louis, MO, USA \\
  Computing and Computational Sciences Directorate \\
  Oak Ridge National Laboratory \\
  Oak Ridge, TN, USA \\
  Lawrence Berkeley National Laboratory \\
  Berkeley, CA, USA \\
  \And 
  \orcidlink{0000-0002-2358-522X} \textbf{Tirthankar Ghosal} \\
  Computer Science and Mathematics Division \\
  Computing and Computational Sciences Directorate \\
  Oak Ridge National Laboratory \\
  Oak Ridge, TN, USA \\
  \And
  \orcidlink{0000-0002-4173-9659} \textbf{Markus J. Buehler}$^{*}$ \\
  Department of Civil and Environmental Engineering \\
  Department of Mechanical Engineering \\
  Schwarzman College of Computing \\
  Massachusetts Institute of Technology \\
  Cambridge, MA, USA \\
  \\
  $^{*}$Corresponding author: \texttt{mbuehler@MIT.EDU}
}

\begin{document}
\maketitle

\begin{abstract}
Accelerating materials discovery requires AI systems that can generate scientifically valid hypotheses through multi-step, domain-grounded reasoning. Standard large language models often produce fluent but weakly traceable responses to open-ended materials design problems, making it difficult to determine whether final answers are supported by coherent intermediate reasoning. We develop Graph-PRefLexOR, a family of graph-native reasoning models fine-tuned with Group Relative Policy Optimization (GRPO) to organize reasoning into explicit phases for mechanism exploration, graph construction, pattern extraction, and hypothesis synthesis. This design links neural language generation with symbolic relational structure, enabling causal connections to be constructed, inspected, and reused. On 100 open-ended questions from materials science and mechanics literature, Graph-PRefLexOR achieves 40-65\% improvements over corresponding base models, with the largest gains in reasoning traceability. Embedding analyses show broader semantic exploration and approximately $2\times$--$3\times$ greater semantic diversity than baselines. Semantic backtracking and layer-wise hidden-state analyses further show stronger alignment between structured reasoning and final answers. Finally, test-time graph expansion reveals that additional compute primarily increases long-range conceptual recombination within a bounded semantic space, rather than simply expanding semantic coverage. These results establish graph-native reinforcement learning as a pathway toward interpretable AI systems for scientific hypothesis generation in materials design and other scientific applications.
\end{abstract}


\section{Introduction}

Scientific discovery increasingly depends on the ability to connect concepts, mechanisms, and evidence across domains that are often studied in isolation. This challenge is especially pronounced in materials science and mechanics, where macroscopic properties emerge from coupled processes spanning molecular structure, mesoscale organization, interfaces, defects, processing history, and boundary conditions~\cite{wang_scientific_2023,buehler_accelerating_2024,nepal_hierarchically_2023,wegst_bioinspired_2015}. Generating useful hypotheses in such settings requires more than retrieving relevant facts or summarizing prior work; it requires organizing relationships among entities, mechanisms, constraints, and outcomes. Yet scientific knowledge remains fragmented across papers, disciplines, terminologies, and modeling frameworks, leaving many potentially important connections implicit or difficult to evaluate~\cite{swanson_undiscovered_1986}. The central problem, therefore, is how to build AI systems that can transform dispersed scientific information into interpretable reasoning structures capable of supporting cross-domain hypothesis generation.

Large language models (LLMs) offer a promising substrate for this task because much of scientific knowledge is encoded in text, and LLMs can summarize literature, compare concepts, generate explanations, and assist in hypothesis formulation~\cite{vaswani_attention_2017,brown_language_2020,zhao_survey_2026,luu_bioinspiredllm_2024,ghafarollahi_protagents_2024}. Recent AI-for-science workflows have extended these capabilities toward materials discovery, protein design, code generation, autonomous experimentation, and scientific agents~\cite{lu_ai_2024,hage_beamperl_2026,buehler_mechgpt_2024,buehler_melm_2023,ghafarollahi_sciagents_2025,ghafarollahi_sparks_2025}. However, most LLM reasoning remains expressed as linear text. Chain-of-thought and related prompting strategies can improve performance by exposing intermediate reasoning, but these traces often lack an explicit representation of the entities, relations, dependencies, and causal links that organize scientific explanations~\cite{wei_chain--thought_2022,yao_tree_nodate,lanham_measuring_2023}. This limitation is consequential for materials reasoning, where hypotheses often require connecting concepts across different domains and understanding failure modes at multiple scales. A textual trace may describe such links, but it does not directly encode which concepts are connected, what type of relation connects them, or how local mechanisms compose into a higher-level explanation.

Prior approaches have begun to address parts of this problem. Retrieval-augmented generation improves access to relevant documents, knowledge graphs encode entities and relations explicitly, and agentic workflows decompose complex scientific tasks across specialized components~\cite{lewis_retrieval-augmented_2020,gao_retrieval-augmented_2024,venugopal_matkg_2024,pan_unifying_2024,wang_autonomous_2026,ghafarollahi_automating_2025,ghafarollahi_rapid_2025}. Graph representations are particularly attractive because they make scientific relationships inspectable: concepts can be represented as nodes, while mechanisms, causal dependencies, analogies, constraints, or failure pathways can be represented as directed edges~\cite{hogan_knowledge_2022,stewart_graphagents_2026}. Nevertheless, existing systems often use graphs as external retrieval substrates or post hoc representations rather than as the model's native reasoning format. Similarly, agentic systems may exchange artifacts or natural-language messages without forcing the underlying reasoning to become relational, parseable, or reusable. Thus, while existing methods improve access, orchestration, and organization of scientific information, they do not fully solve the problem of making the model's intermediate reasoning itself graph-structured and evaluable.

Here, we address this gap with Graph-PRefLexOR, a graph-native reasoning model for open-ended scientific hypothesis generation~\cite{buehler_preflexor_2025,buehler_situ_2025}. Building on earlier Graph-PRefLexOR formulations that mapped a task to a knowledge graph, abstract patterns, and a final answer, the present model exposes a more granular sentinel-based reasoning trace consisting of \texttt{<brainstorm>}, \texttt{<graph>}, \texttt{<graph\_json>}, \texttt{<patterns>}, and \texttt{<synthesis>} phases. These stages separate mechanism exploration, concept abstraction, machine-readable graph construction, higher-order pattern identification, and final hypothesis synthesis. We train the model using Group Relative Policy Optimization (GRPO) \cite{shao_deepseekmath_2024}, a reinforcement-learning method that improves model behavior through relative comparisons among groups of generated outputs rather than relying on fixed target answers alone. In this setting, GRPO encourages the model to generate reasoning traces that are not merely fluent but also structured, relational, and traceable. This differs from our earlier ORPO/EXO-style preference-optimization approaches by directly optimizing the model toward higher-quality reasoning behavior through group-relative rewards, making it well suited for open-ended scientific problems where multiple plausible hypotheses may exist.

We evaluate Graph-PRefLexOR on a manually curated benchmark of 100 open-ended questions derived from materials science and mechanics literature. The benchmark probes cross-domain linkage, causal mapping, hidden-variable identification, model abstraction, and hypothesis generation capabilities that are difficult to assess using standard factual or multiple-choice evaluations. Across 1.7B, 3B, and 8B model scales, Graph-PRefLexOR outperforms its corresponding base models in reasoning quality, intellectual depth, and reasoning traceability, with the strongest improvements observed in traceability. Embedding-based analyses further show that Graph-PRefLexOR reasoning traces occupy more organized semantic regions, follow more directional trajectories, and exhibit approximately $2\times$--$3\times$ greater semantic diversity than baseline traces. Semantic backtracking and layer-wise hidden-state analyses show that final answers remain strongly aligned with the structured reasoning pathway, particularly the \texttt{<synthesis>} stage, indicating improved continuity between intermediate reasoning and response generation. We further show that the graph-native format supports test-time graph expansion: by accumulating emitted graph structures into a growing memory graph, additional inference-time compute produces statistically novel long-range conceptual recombinations within a bounded semantic space. Together, these results indicate that graph-native GRPO improves scientific reasoning not simply by changing final answers, but by restructuring the intermediate computational pathway through which hypotheses are generated, aligned, and iteratively recombined.

\begin{figure}
    \centering
    \includegraphics[width=1\linewidth]{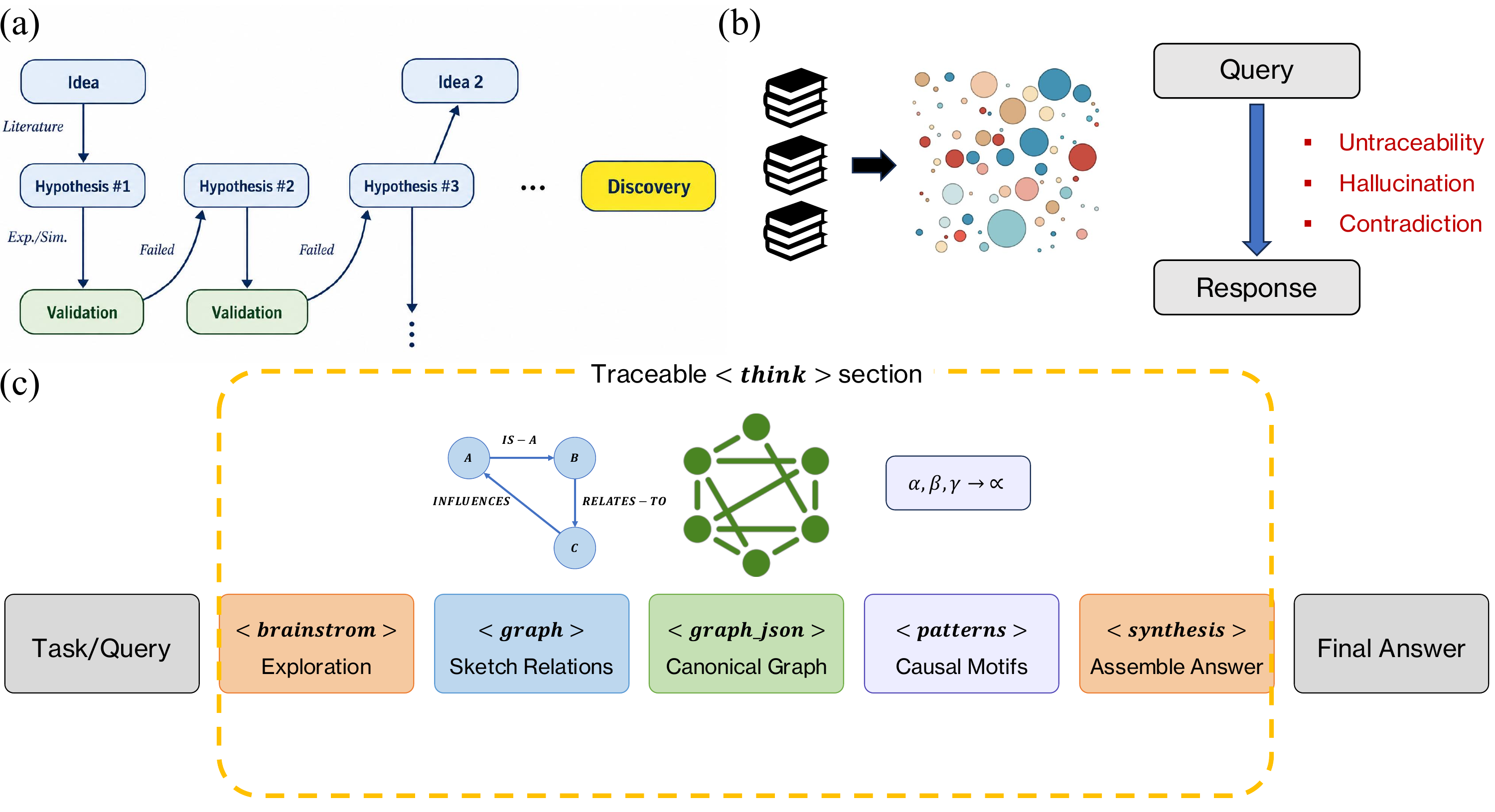}
    \caption{(a) Scientific discovery often proceeds through iterative hypothesis generation, validation, re-ideation and refinement. (b) Standard LLM responses to scientific queries can be difficult to trace, leading to untraceability, hallucination, or contradiction. (c) Graph-PRefLexOR addresses this limitation by organizing the \texttt{<think>} section into explicit reasoning phases: \texttt{<brainstorm>} for mechanism exploration, \texttt{<graph>} for relation sketching, \texttt{<graph\_json>} for canonical graph construction, \texttt{<patterns>} for causal motif extraction, and \texttt{<synthesis>} for assembling the final answer.}
    \label{fig:cover_photo}
\end{figure}

\section{Results and Discussion}

\subsection{Evaluation of Structured Reasoning}

One of the primary objectives of this study is to quantitatively assess the quality of reasoning traces generated by the graph-native Group Relative Policy Optimization (GRPO) model compared to its corresponding base models \cite{buehler_situ_2025, buehler_preflexor_2025, shao_deepseekmath_2024}. To this end, we develop three GRPO-based variants, collectively termed Graph-PRefLexOR, at scales of 8B, 3B, and 1.7B parameters. These models are initialized from Qwen3-8B \cite{yang_qwen3_2025}, Llama-3.2-3B-Instruct \cite{noauthor_llama_nodate}, and Qwen3-1.7B \cite{yang_qwen3_2025}, respectively, and subsequently fine-tuned to perform structured, multi-stage reasoning.

Specifically, the models are trained to organize their internal reasoning into a sequence of structured phases within the thinking process: \texttt{<brainstorm>}, \texttt{<graph>}, \texttt{<graph\_json>}, \texttt{<patterns>}, and \texttt{<synthesis>}. The \texttt{<brainstorm>} phase performs divergent exploration, generating candidate mechanisms, hypotheses, and potential failure modes grounded in domain knowledge. The \texttt{<graph>} phase then abstracts this reasoning into a conceptual representation by identifying core entities (e.g., sequence, structure, processing, properties) and their causal relationships, forming an interpretable reasoning scaffold. This abstraction is formalized in the \texttt{<graph\_json>} phase as a machine-readable directed graph, where nodes correspond to entities and edges encode typed relationships (e.g., \texttt{"source": "Sequence", "relation": "encodes", "target": "Structure"}), drawn from a predefined relational vocabulary \cite{hogan_knowledge_2022}. The \texttt{<patterns>} phase subsequently extracts higher-order regularities from this graph, such as causal chains (e.g., sequence $\rightarrow$ structure $\rightarrow$ properties $\rightarrow$ failure), scale-bridging relationships, and feedback loops that capture recurring mechanistic motifs. Finally, the \texttt{<synthesis>} phase integrates these patterns into a coherent, testable hypothesis, explicitly linking multi-scale mechanisms to predicted outcomes and potential failure modes. 

To evaluate the performance gains achieved by the GRPO-based models, we construct an open-ended benchmark comprising 100 questions designed to probe cross-disciplinary linkage, causal mapping, and hypothesis generation capabilities, derived from published materials science and mechanics literature. Additional details regarding dataset construction are provided in the Methods section. We perform inference across six models with reasoning-enabled decoding (except for the Llama-3.2-3B-Instruct baseline), allowing each model to explicitly generate intermediate reasoning during the \texttt{<think>} phase.

Because the final responses are conditioned on this internal reasoning process, we evaluate the quality of the generated \texttt{<think>} traces directly. For this purpose, we define three complementary evaluation metrics: \textit{Reasoning Quality}, \textit{Intellectual Depth}, and \textit{Reasoning Traceability}, along with an aggregate \textit{Overall Score} computed as their mean. Each metric is scored on a 0--10 scale, and evaluation is performed using Claude \emph{opus-4.7} as an independent judge \cite{zheng_judging_2023, noauthor_introducing_nodate}. Notably, as OpenAI GPT models are employed during dataset generation, we use Claude-based evaluation to mitigate potential model-family bias. Collectively, these metrics capture distinct dimensions of reasoning performance, enabling a structured assessment of both the correctness and organization of the generated reasoning traces.

\begin{figure}
    \centering
    \includegraphics[width=1\linewidth]{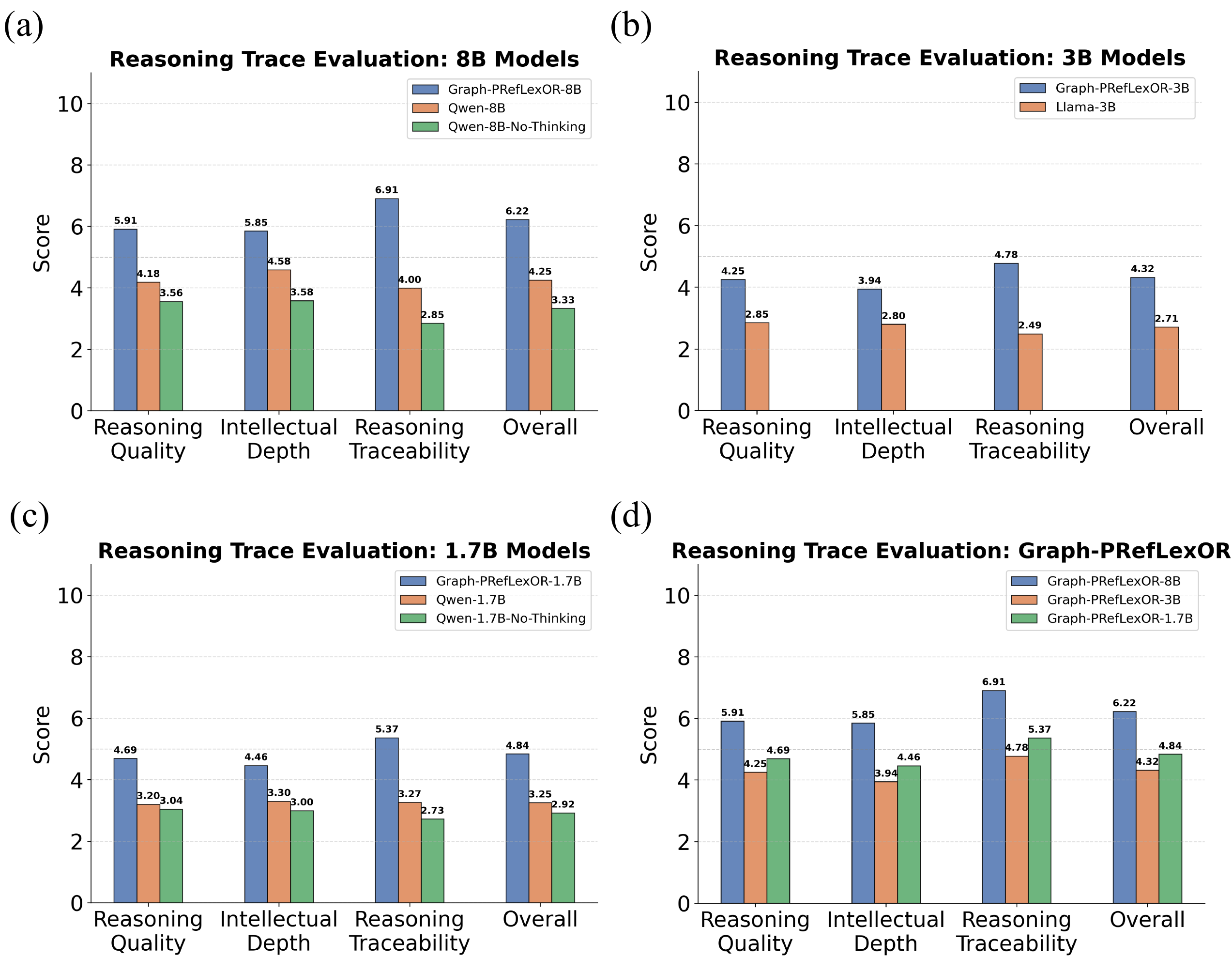}
    \caption{Evaluation of structured reasoning across model scales on open-ended scientific questions ($N=100$), assessed using Claude Opus-4.7. Metrics include Reasoning Quality, Intellectual Depth, Reasoning Traceability, and Overall score (0–10). (a) Graph-PRefLexOR-8B vs. Qwen3-8B (with no-thinking variant), (b) Graph-PRefLexOR-3B vs. Llama-3.2-3B-Instruct, and (c) Graph-PRefLexOR-1.7B vs. Qwen3-1.7B (with no-thinking variant). Across all scales, Graph-PRefLexOR consistently outperforms its base models on all metrics, with the largest gains in reasoning traceability, indicating improved structural organization and causal transparency in hypothesis generation from research articles.}
    \label{fig:reasoning_model}
\end{figure}

Figure \ref{fig:reasoning_model} summarizes the performance of the Graph-PRefLexOR models relative to their corresponding base models across the three evaluation metrics. Across all model scales, the GRPO-trained graph models consistently outperform their baselines, exhibiting improvements of approximately 40--65\% in aggregate performance, with the largest gains observed in \textit{Reasoning Traceability}. This trend indicates that structured, graph-native reasoning primarily enhances the model's ability to construct mechanistic and causally grounded explanations, rather than merely improving surface-level coherence. Moreover, summarizing core concepts using a directed knowledge graph further enriches reasoning quality \cite{hogan_knowledge_2022}.

For the Llama-3.2-3B-Instruct baseline, evaluation is restricted to the final response, as the model lacks an explicit reasoning phase. To further isolate the contribution of structured reasoning, we additionally evaluate Qwen-based models with reasoning disabled (\emph{no-thinking} setting). The resulting performance degradation closely mirrors that observed for the Llama baseline, with overall reductions on the order of 30--50\%, confirming that the majority of the observed gains arise from the explicit reasoning process rather than architectural differences alone.

A second key trend emerges with respect to model scale and base model characteristics. As shown in Fig.~\ref{fig:reasoning_model}d, the 8B Graph-PRefLexOR model consistently outperforms its smaller counterparts, achieving approximately 25--30\% higher scores than the 1.7B variant across all metrics. This improvement is primarily attributable to increased representational capacity, which enables more expressive and coherent graph construction and pattern extraction. Furthermore, the 3B model, initialized from Llama-3.2-3B-Instruct, underperforms relative to the 1.7B Qwen-based model. This behavior suggests that the absence of an inherent reasoning scaffold in the base model limits the effectiveness of GRPO-based structured reasoning.

\subsection{Qualitative Comparison of Reasoning Traces}
\label{sec:qualitative_comparison}
\begin{figure}
    \centering
    \includegraphics[width=1\linewidth]{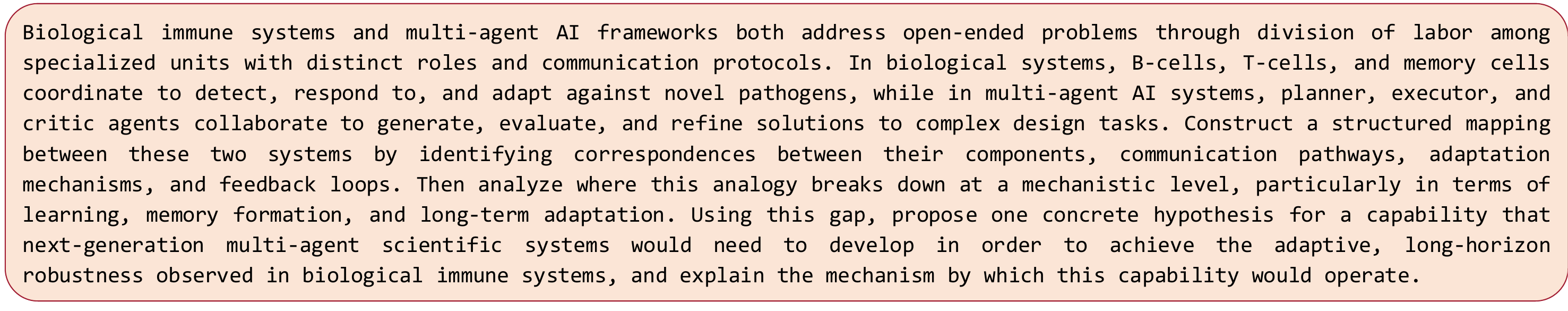}
   \caption{Representative cross-disciplinary hypothesis-generation question used to evaluate Graph-PRefLexOR. The question is derived from Ref.~\cite{ghafarollahi_protagents_2024} and probes analogical mapping, mechanistic breakdown, and long-horizon adaptive reasoning.}
    \label{fig:question}
\end{figure}

To provide a qualitative comparison of model behavior, we examine representative responses from Graph-PRefLexOR-8B and Qwen3-8B to the same benchmark question. Figure~\ref{fig:question} shows an example question used to evaluate open-ended reasoning quality. The question is derived from the multi-agent protein-design framework of Ghafarollahi and Buehler~\cite{ghafarollahi_protagents_2024}, but is written to be self-contained and does not require direct knowledge of the original paper. The question asks the model to construct an analogy between two superficially distinct hierarchical systems: biological immune response and multi-agent AI coordination. Specifically, it requires the model to identify correspondences among specialized components, communication pathways, adaptation mechanisms, and feedback loops. It then asks the model to analyze where the analogy breaks down mechanistically, particularly with respect to learning, memory formation, and long-term adaptation, and to use this gap to propose a concrete future capability for next-generation multi-agent scientific systems. This structure makes the question a useful probe of cross-domain mapping, mechanistic reasoning, limitation analysis, and hypothesis generation.

\begin{figure}
    \centering
    \includegraphics[width=1\linewidth]{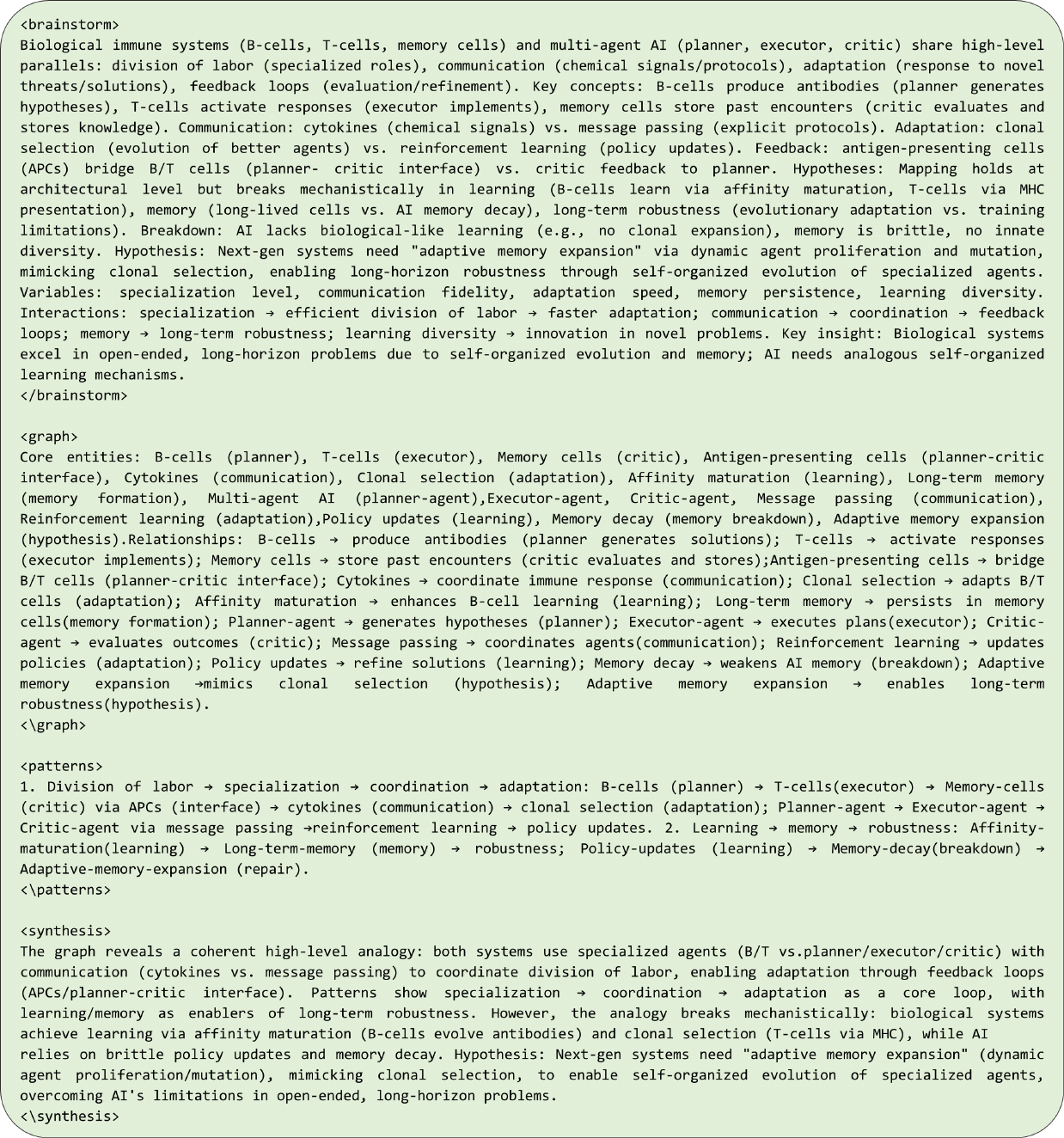}
    \caption{Representative Graph-PRefLexOR-8B reasoning to the benchmark question in Fig. \ref{fig:question}. The response is organized into structured reasoning phases, including \texttt{<brainstorm>}, \texttt{<graph>}, \texttt{<patterns>}, and \texttt{<synthesis>}. This format exposes the intermediate pathway from analogical mapping to mechanistic gap identification and final hypothesis generation.}
    \label{fig:graph_ans}
\end{figure}

Figure~\ref{fig:graph_ans} shows the corresponding Graph-PRefLexOR-8B reasoning trace. The model decomposes its reasoning into explicit phases, beginning with divergent mechanism exploration in \texttt{<brainstorm>}, followed by entity--relation abstraction in \texttt{<graph>}, higher-order pattern extraction in \texttt{<patterns>}, and final hypothesis integration in \texttt{<synthesis>}. This structure makes the reasoning trace directly inspectable: the analogy is first explored conceptually, then converted into mapped components and relations, and finally synthesized into a mechanistic hypothesis. In this example, the model identifies \emph{adaptive memory expansion} as a candidate capability for long-horizon robustness in multi-agent AI systems, motivated by biological mechanisms such as clonal selection, affinity maturation, and persistent immune memory.

\begin{figure}
    \centering
    \includegraphics[width=0.8\linewidth]{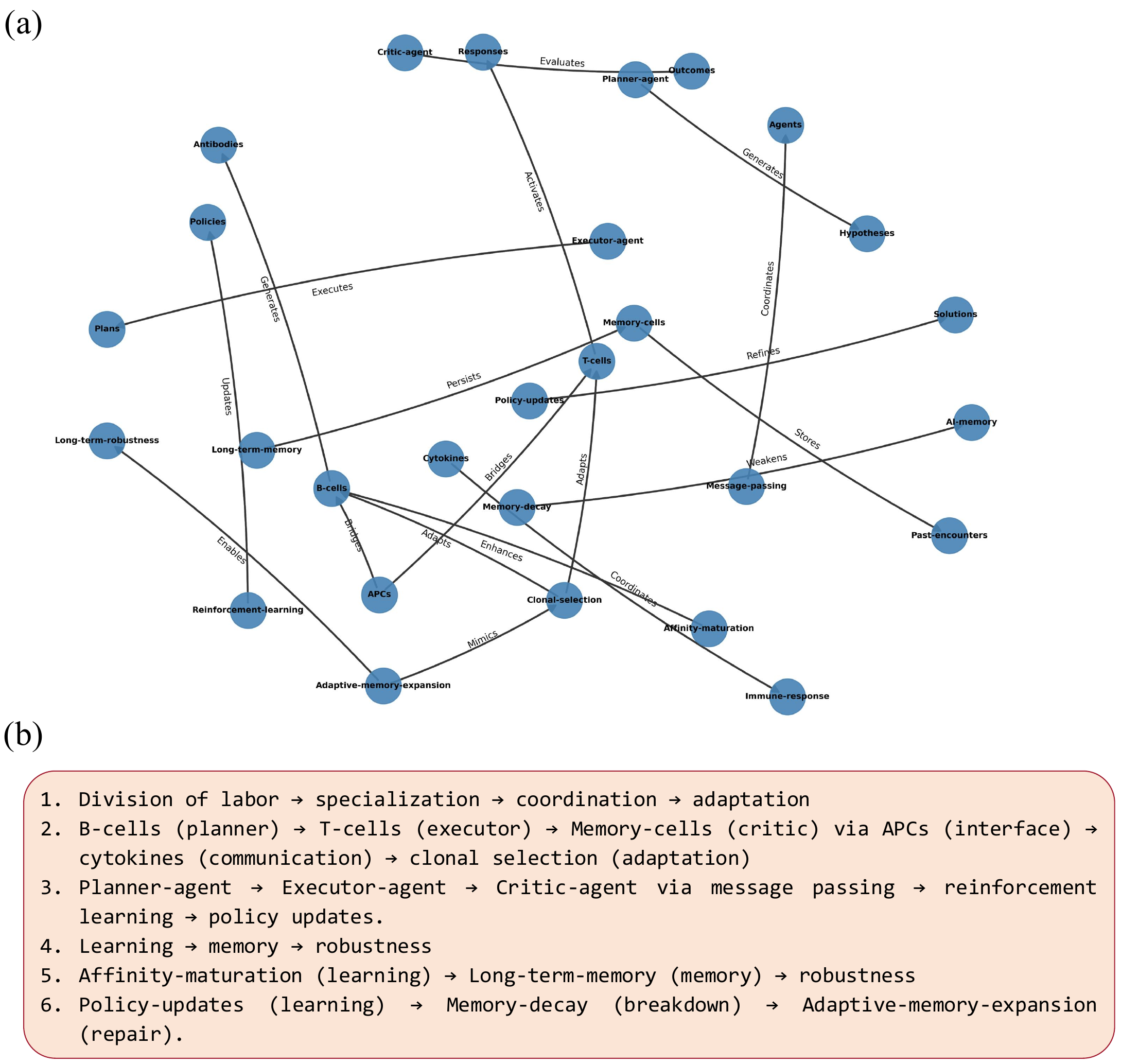}
    \caption{Graph and pattern representation extracted from the Graph-PRefLexOR-8B response. (a) Directed graph linking biological immune-system concepts, multi-agent AI components, and the proposed bridging mechanism. (b) Higher-order reasoning patterns extracted from the graph, summarizing the main causal motifs used for hypothesis synthesis.}
    \label{fig:graph_pattern}
\end{figure}

The graph and pattern representations extracted from the same response are shown in Fig.~\ref{fig:graph_pattern}. The directed graph encodes key immune-system concepts, multi-agent AI components, and proposed bridging mechanisms as nodes connected by typed relations. Biological immune-system concepts are grouped separately from multi-agent AI concepts, while the proposed bridging mechanism links adaptive memory expansion to clonal selection and long-term robustness. The pattern panel further compresses the graph into higher-order motifs, such as division of labor $\rightarrow$ specialization $\rightarrow$ coordination $\rightarrow$ adaptation and learning $\rightarrow$ memory $\rightarrow$ robustness. These motifs reveal how Graph-PRefLexOR converts a long-form reasoning trace into a compact relational structure that can be inspected and reused.

In contrast, the Qwen3-8B baseline produces a substantially longer linear response that contains relevant concepts but lacks an explicit relational scaffold. Because the Qwen3-8B response is substantially longer than the Graph-PRefLexOR trace, only its qualitative behavior is summarized in the main text; the full baseline response is provided in Supplementary Information. The response proceeds through extended prose, repeated uncertainty, and backtracking before arriving at a final conclusion. While such a response can still identify useful factors, its intermediate reasoning is less compact, less parseable, and less directly organized around entities, relations, and reusable causal motifs. This qualitative comparison motivates the embedding-based analyses that follow, where we quantify whether these visible structural differences correspond to measurable changes in semantic organization, trajectory directionality, and reasoning diversity.

\subsection{Geometry of Reasoning Representations}
\subsubsection{Semantic Organization}
To further characterize the semantic structure of the generated reasoning traces, we project both intermediate reasoning steps and final answers into a shared embedding space using the \texttt{google/embeddinggemma\_300m} model, which maps text to 768-dimensional vectors \cite{vera_embeddinggemma_2025}. For each model, we embed both the reasoning traces (i.e., \texttt{<think>} content) and the corresponding final answers. The reasoning traces are segmented into atomic reasoning steps, where each step is defined as a complete sentence without paragraph breaks. This results in an $n \times 768$ representation for each sample, where $n$ denotes the number of reasoning steps.

We then apply Principal Component Analysis (PCA) to project these embeddings into a two-dimensional space for visualization \cite{abdi_principal_2010}. A key observation is that baseline models consistently generate longer reasoning traces, resulting in a larger number of reasoning steps compared to the GRPO-trained models. Direct point-wise visualization would therefore bias the comparison toward baseline models due to their higher sampling density. To address this, we estimate the underlying distribution of reasoning steps using Gaussian kernel density estimation (KDE) and visualize the resulting density as contour plots over a shared embedding grid \cite{scott_multivariate_2012}. This enables a distributional comparison that is invariant to the number of reasoning steps.

For reasoning trace analysis, we focus only on the 8B and 1.7B models to isolate the effects of scale while maintaining comparable base architectures. The GRPO-generated reasoning steps are further partitioned according to their structured phases: \texttt{<brainstorm>}, \texttt{<graph>}, \texttt{<patterns>}, and \texttt{<synthesis>}, and KDE is applied separately to each phase to reveal their semantic organization. In contrast, baseline models are represented as a single aggregated distribution. For final answer analysis, we apply the same embedding and KDE procedure across all three model scales, treating each answer as a single semantic distribution without phase decomposition.

\begin{figure}
    \centering
    \includegraphics[width=1\linewidth]{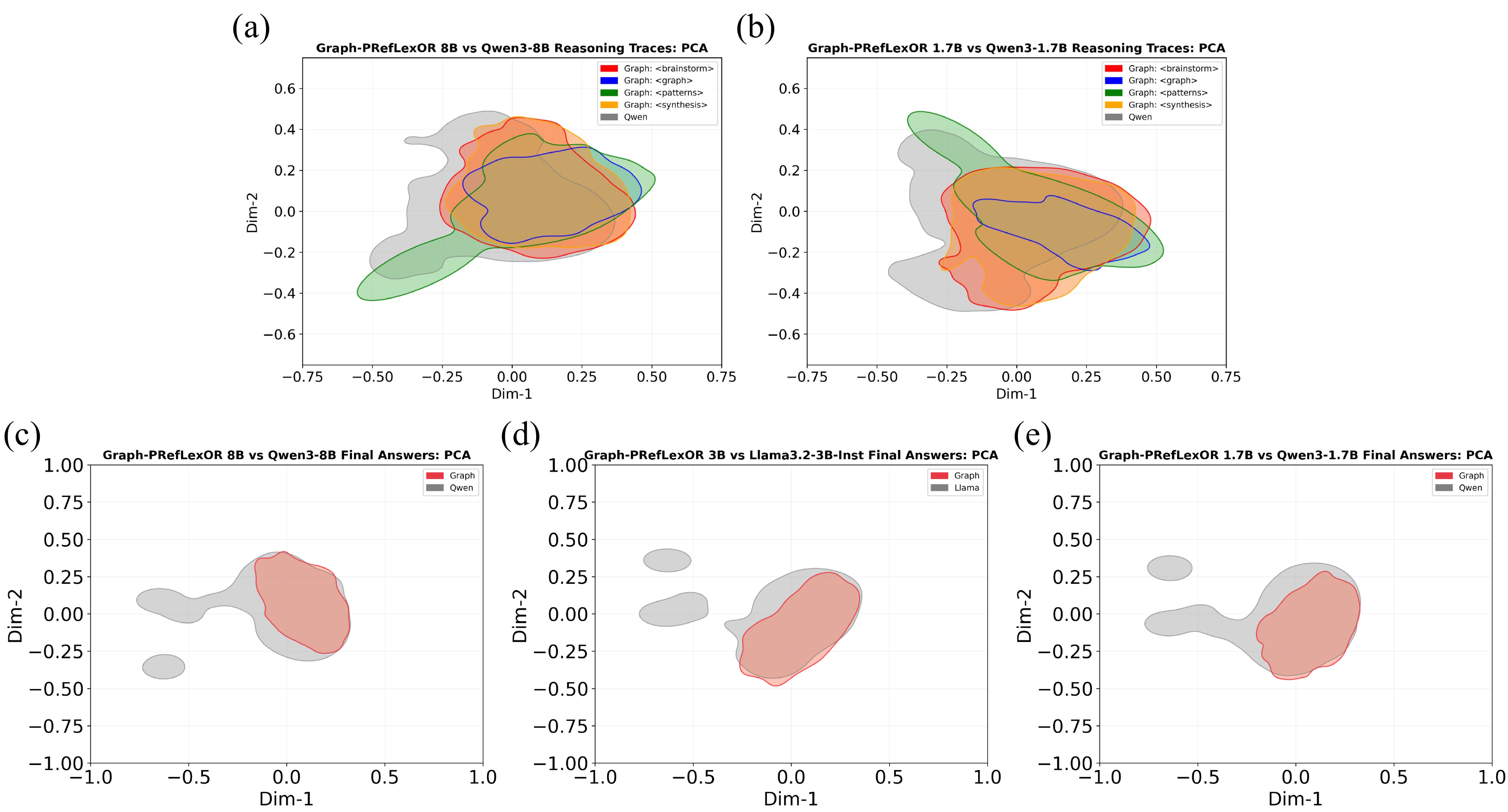}
    \caption{PCA projection of reasoning traces and final answers comparing Graph-PRefLexOR and base models across scales. (a) Graph-PRefLexOR-8B vs. Qwen3-8B reasoning traces, (b) Graph-PRefLexOR-1.7B vs. Qwen3-1.7B reasoning traces, and (c–e) corresponding comparisons of final answers for 8B, 3B, and 1.7B models, respectively. Reasoning traces are decomposed into structured components (\texttt{<brainstorm>}, \texttt{<graph>}, \texttt{<patterns>}, and \texttt{<synthesis>}), revealing more organized and separable distributions relative to baseline models. In contrast, final answer representations exhibit tighter clustering and greater alignment between Graph-PRefLexOR and base models, indicating that performance gains arise primarily from improved intermediate reasoning structure rather than differences in final outputs.}
    \label{fig:semantic_pca}
\end{figure}

Figure \ref{fig:semantic_pca} reveals several key trends in the semantic organization of reasoning traces and final outputs. First, the reasoning traces of Graph-PRefLexOR exhibit a more structured, separable distribution across the embedding space than those of baseline models. The phase-wise decomposition yields distinct yet partially overlapping regions, indicating that each phase occupies a specialized semantic subspace while contributing to a coherent reasoning trajectory. In contrast, baseline models form a single, diffuse distribution.

Second, the Graph-PRefLexOR distributions span a broader and more directional manifold, particularly evident in the \texttt{<brainstorm>} and \texttt{<patterns>} phases, which extend into regions not covered by the baseline models. This suggests enhanced exploration of the hypothesis space and the emergence of higher-order abstractions that are not captured by standard autoregressive reasoning.

Third, despite these substantial differences in intermediate reasoning representations, the final answer embeddings exhibit significantly tighter clustering and strong overlap between Graph-PRefLexOR and baseline models across all scales (Fig. \ref{fig:semantic_pca}c--e). This convergence indicates that both model classes converge on semantically similar endpoints, even though their underlying reasoning trajectories differ markedly. However, the baseline final answer embeddings exhibit multiple separated clusters in the PCA space, suggesting fragmentation in semantic representations across samples. In contrast, Graph-PRefLexOR produces a more compact and unified distribution, indicating greater consistency in the final answer space. Collectively, these observations support the hypothesis that the primary advantage of Graph-PRefLexOR lies in the organization and expressivity of intermediate reasoning, rather than in substantial shifts in the final answer distribution.

\subsubsection{Directed Trajectories}
The PCA analysis indicates that Graph-PRefLexOR occupies broader and more structured regions of the latent embedding space than the corresponding base models. However, this distributional view captures only where the representations lie, not how the model moves through semantic space during reasoning and answer generation. To examine this dynamic aspect, we represent each response as an ordered trajectory in the PCA-projected embedding space. For Graph-PRefLexOR, reasoning trajectories are constructed from the four structured reasoning phases: \texttt{<brainstorm>}, \texttt{<graph>}, \texttt{<patterns>}, and \texttt{<synthesis>}. Final answers are instead divided into four equal-length sequential chunks. For baseline models, which lack explicit phase annotations, both reasoning traces and final answers are partitioned into four equal-length sequential chunks. Consecutive phases or chunks are then connected by directed arrows, enabling visualization of how each model traverses semantic space over the course of reasoning and response generation.

\begin{figure}
    \centering
    \includegraphics[width=0.8\linewidth]{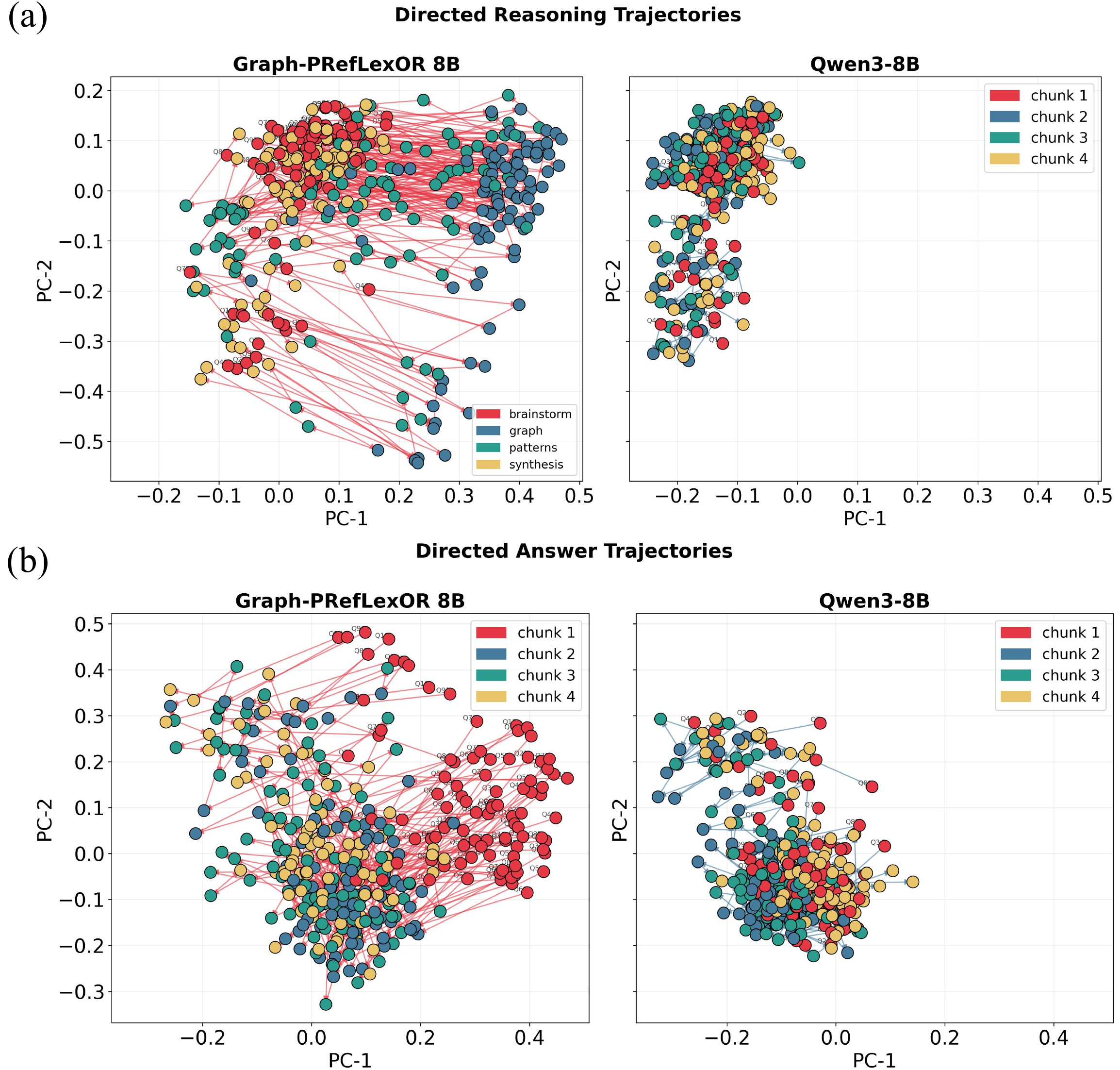}
    \caption{PCA projection of directed (a) reasoning, and (b) answer trajectories between Graph-PRefLexOR-8B and Qwen3-8B. For Graph-PRefLexOR, trajectories explicitly follow structured stages (\texttt{<brainstorm>}, \texttt{<graph>}, \texttt{<patterns>}, and \texttt{<synthesis>}), forming coherent, directional transitions in latent space. In contrast, base model trajectories (shown as sequential chunks) remain more localized and less structured. For answer trajectories, four sequential chunks are used for both models.}
    \label{fig:pca_traj}
\end{figure}

This formulation enables a direct comparison between structured and unstructured reasoning dynamics in latent space. In Graph-PRefLexOR, trajectories correspond to transitions between functionally distinct reasoning operations, revealing how the model progressively reorganizes semantic representations throughout the reasoning process. As shown in Fig.~\ref{fig:pca_traj}a, the trajectory initially shifts from the \texttt{<brainstorm>} phase toward the \texttt{<graph>} phase, reflecting a transition from divergent hypothesis generation to the abstraction of key entities and their relationships. The subsequent \texttt{<patterns>} phase remains closer to the graph representation, consistent with its role in identifying higher-order causal and symbolic relationships among the extracted concepts. Finally, the \texttt{<synthesis>} phase integrates the hypotheses, concepts, and patterns into a coherent mechanistic explanation and partially returns toward the semantic region occupied by \texttt{<brainstorm>}, suggesting that synthesis reconnects structured abstractions with the original hypothesis space. Collectively, these transitions produce broad, directional movements through the embedding space, indicating a richer and more organized reasoning process.

In contrast, Qwen3-8B trajectories remain comparatively localized and entangled, with sequential chunks exhibiting substantial overlap and limited directional separation. This behavior is consistent with unstructured reasoning patterns such as backtracking, self-correction, and repetition, which compress the trajectories into narrower regions of latent space. Consequently, baseline trajectories resemble generic sequential continuation through generated text rather than transitions between functionally distinct reasoning states.

For answer trajectories (Fig.~\ref{fig:pca_traj}b), a similar but less pronounced trend is observed. Because final answers are divided into four equal, sequential chunks in both models, the trajectories reflect the evolution of the semantic content of the response during answer generation. Graph-PRefLexOR exhibits a broader early displacement, with the second chunk moving furthest from the first, followed by subsequent chunks that remain comparatively closer in semantic space. This suggests that the final answer initially expands or reframes the response before consolidating into a more stable explanatory structure.

In contrast, Qwen3-8B trajectories remain more compact throughout the answer-generation process, with stronger overlap among sequential chunks and limited directional separation. Thus, although both models show greater convergence in final answer space than in reasoning-trace space, Graph-PRefLexOR retains a more pronounced pattern of early semantic diversification followed by synthesis. Overall, these observations reinforce that Graph-PRefLexOR induces more structured and directional semantic evolution, whereas the baseline model exhibits comparatively localized and less differentiated trajectories.

\subsection{Quantifying Semantic Diversity}
The trajectory analysis provides qualitative evidence that Graph-PRefLexOR generates richer and more differentiated reasoning representations. To complement this visualization with a quantitative measure, we compute centroid-based semantic distances in the original 768-dimensional embedding space. For reasoning traces, Graph-PRefLexOR embeddings are grouped according to their structured phases. For each response, we compute the centroid of each phase and evaluate all six pairwise cosine distances between phase centroids. These distances are then averaged to obtain a single response-level measure of semantic diversity across reasoning stages.

For baseline models, which lack explicit phase annotations, each reasoning trace is partitioned into four approximately equal sequential chunks and treated as pseudo-phases as before. The same centroid-distance calculation is applied, enabling a matched comparison between structured and unstructured reasoning. This analysis is performed for the 1.7B and 8B model pairs. We apply an analogous procedure to final answers across all model scales (1.7B, 3B, and 8B), dividing each answer into four equal sequential chunks for both Graph-PRefLexOR and the corresponding baseline model. The resulting distributions are visualized using violin plots with jittered samples, median indicators, and mean $\pm$ standard deviation overlays.

\begin{figure}
    \centering
    \includegraphics[width=0.8\linewidth]{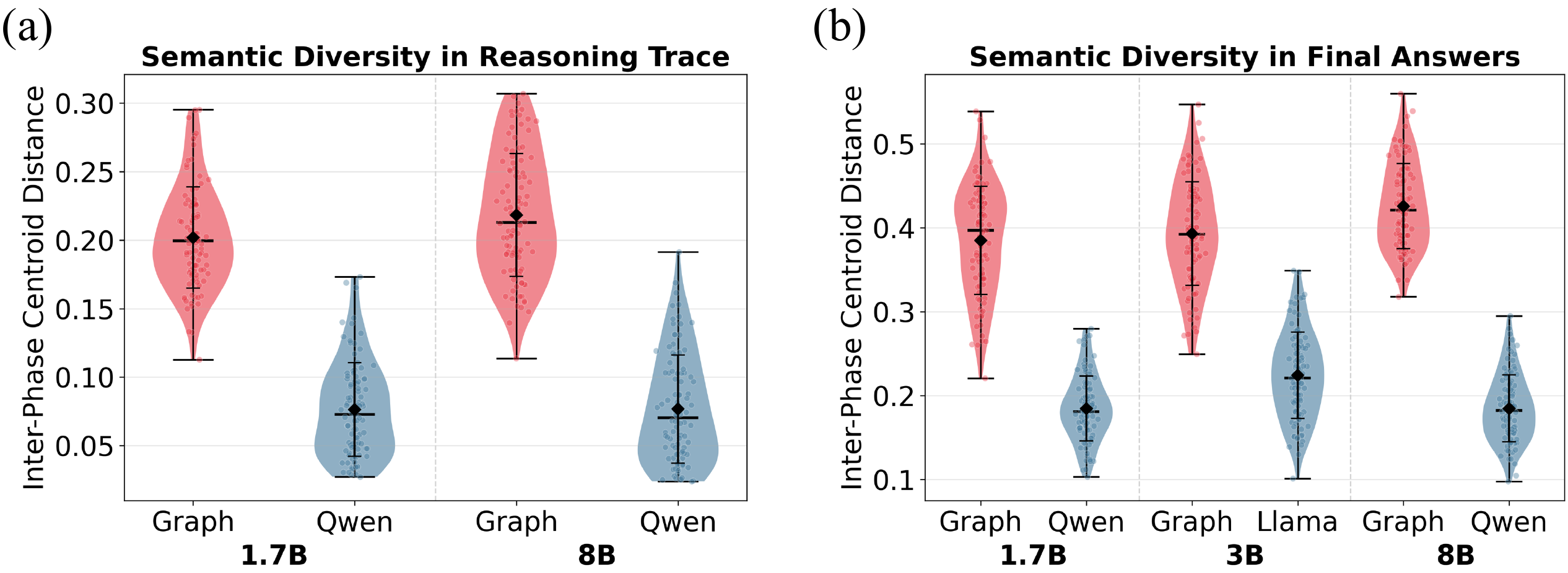}
    \caption{Semantic diversity measured via inter-phase centroid distance for (a) reasoning traces and (b) final answers across model scales. Violin plots show the distribution of sample-level semantic diversity scores for Graph-PRefLexOR and the corresponding base models. Individual points denote responses, horizontal black lines indicate medians, black diamonds denote means, and vertical error bars represent one standard deviation. Graph-PRefLexOR consistently exhibits higher inter-phase distances across all scales, indicating greater semantic separation between reasoning stages and increased diversity in generated representations. This effect is more pronounced in reasoning traces and remains evident, though reduced, in final answers.}
    \label{fig:semantic_diversity}
\end{figure}

Figure~\ref{fig:semantic_diversity} shows that Graph-PRefLexOR consistently exhibits higher semantic diversity than its corresponding base models across both reasoning traces and final answers. For reasoning traces (Fig.~\ref{fig:semantic_diversity}a), the mean inter-phase cosine distance increases from 0.07 to 0.20 at 1.7B and from 0.08 to 0.21 at 8B, corresponding to $2.9\times$ and $2.6\times$ gains, respectively (Table~\ref{tab:semantic_diversity_gain}). These larger distances indicate stronger semantic differentiation among reasoning stages, consistent with the structured decomposition of reasoning into distinct functional phases.

\begin{table}[t]
\centering
\caption{Mean semantic diversity and relative gains of Graph-PRefLexOR over corresponding base models.}
\label{tab:semantic_diversity_gain}
\begin{tabular}{llcccc}
\hline
\textbf{Output Type} & \textbf{Model Scale} & \textbf{Graph-PRefLexOR} & \textbf{Base Model} & \textbf{Absolute Increase} & \textbf{Gain} \\
\hline
Reasoning trace & 1.7B & 0.20 & 0.07 & 0.13 & $2.9\times$ \\
Reasoning trace & 8B   & 0.21 & 0.08 & 0.13 & $2.6\times$ \\
\hline
Final answer & 1.7B & 0.39 & 0.18 & 0.21 & $2.2\times$ \\
Final answer & 3B   & 0.39 & 0.22 & 0.17 & $1.8\times$ \\
Final answer & 8B   & 0.43 & 0.18 & 0.25 & $2.4\times$ \\
\hline
\end{tabular}
\end{table}

For final answers (Fig.~\ref{fig:semantic_diversity}b), Graph-PRefLexOR again maintains higher semantic diversity, although the relative separation is smaller than in the reasoning traces. The mean inter-chunk distance increases from 0.18 to 0.39 at 1.7B, from 0.22 to 0.39 at 3B, and from 0.18 to 0.43 at 8B, corresponding to $2.2\times$, $1.8\times$, and $2.4\times$ gains, respectively (Table~\ref{tab:semantic_diversity_gain}). Thus, Graph-PRefLexOR induces approximately 2-3$\times$ greater semantic improvement across model scales, with the strongest effect occurring during intermediate reasoning. This indicates structured reasoning broadens semantic exploration during computation while partially converging during final answer synthesis.

\subsection{Semantic Backtracking}
\label{sec:answer_backtracking}

The preceding analyses show that Graph-PRefLexOR produces more structured and semantically differentiated reasoning traces than the corresponding base model. We next examine whether the final answer remains aligned with the model's own intermediate reasoning. This question is related to prior work on chain-of-thought faithfulness, which asks whether a model's stated reasoning actually supports its final answer~\cite{wei_chain--thought_2022,lanham_measuring_2023,2024arXiv240213950P}. More broadly, it follows the view that model outputs should be assessed by both final-answer quality, and the extent to which they are supported by appropriate intermediate evidence or explanations~\cite{2017arXiv170303717S}. To test this alignment, we perform a semantic backtracking analysis on the same 100-question benchmark. For each response, we embed the final answer and compare it against candidate reference texts using cosine similarity. The reference with the highest similarity is treated as the closest semantic source of the final answer. For Qwen3-8B, the candidate sources are its own \texttt{<think>} trace, Graph-PRefLexOR-8B's final answer, and Graph-PRefLexOR-8B's \texttt{<brainstorm>}, \texttt{<graph>}, \texttt{<patterns>}, and \texttt{<synthesis>} phases. For Graph-PRefLexOR-8B, we perform the symmetric comparison using its own structured reasoning stages, Qwen3-8B's thinking trace, and Qwen3-8B's final answer.

\begin{figure}[!ht]
    \centering
    \includegraphics[width=\linewidth]{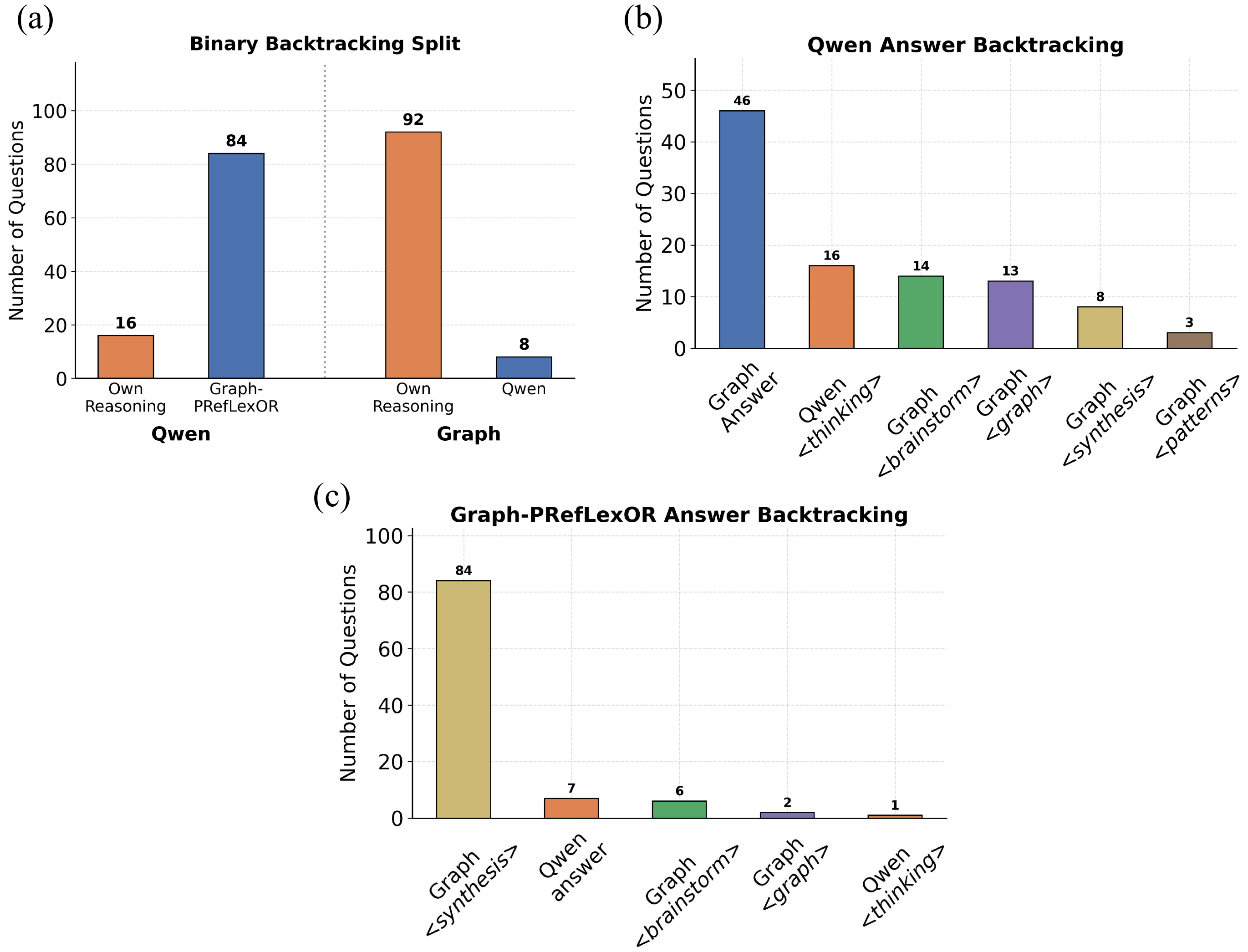}
    \caption{Semantic backtracking analysis of final answer alignment for Qwen3-8B and Graph-PRefLexOR 8B across 100 open-ended scientific questions. (a) Binary split showing whether each final answer is closest to its own reasoning trace or to the other model's outputs. (b) Source distribution for Qwen3-8B final answers, which align with its own \texttt{<think>} trace in only 16/100 cases and more often align with Graph-PRefLexOR outputs. (c) Source distribution for Graph-PRefLexOR-8B final answers, which align with its own structured reasoning stages in 92/100 cases, predominantly the \texttt{<synthesis>} stage.}
    \label{fig:graph_cross_backtracking}
\end{figure}

Figure~\ref{fig:graph_cross_backtracking}a shows a clear asymmetry between the two models. Qwen3-8B final answers are closest to their own thinking traces in only 16 of 100 cases. In the remaining cases, they align more closely with Graph-PRefLexOR-8B-derived outputs, most frequently with Graph-PRefLexOR-8B's final answer itself, which is the closest reference in 46 cases. Additional cases align with Graph-PRefLexOR-8B's \texttt{<brainstorm>} phase in 14 cases, \texttt{<graph>} phase in 13 cases, \texttt{<synthesis>} phase in 8 cases, and \texttt{<patterns>} phase in 3 cases (See Figure \ref{fig:graph_cross_backtracking}b). This indicates that Qwen3-8B can produce final answers that occupy a semantic region similar to Graph-PRefLexOR-8B outputs, even when its own visible reasoning trace is not the closest semantic precursor. In contrast, Graph-PRefLexOR-8B final answers remain strongly anchored to its own structured reasoning pathway. In the cross-model comparison, Graph-PRefLexOR-8B final answers are closest to one of its own reasoning phases in 92 of 100 cases. The dominant source is \texttt{<synthesis>}, which is closest in 84 cases, whereas only 8 cases align more closely with Qwen3-8B outputs (See Figure \ref{fig:graph_cross_backtracking}c). This suggests that Graph-PRefLexOR-8B maintains a tighter semantic connection between intermediate reasoning and final-answer generation. To further isolate the internal structure of this alignment, we compare each Graph-PRefLexOR-8B final answer only against its own reasoning stages: \texttt{<brainstorm>}, \texttt{<graph>}, \texttt{<patterns>}, and \texttt{<synthesis>}.

\begin{figure}[!ht]
    \centering
    \includegraphics[width=\linewidth]{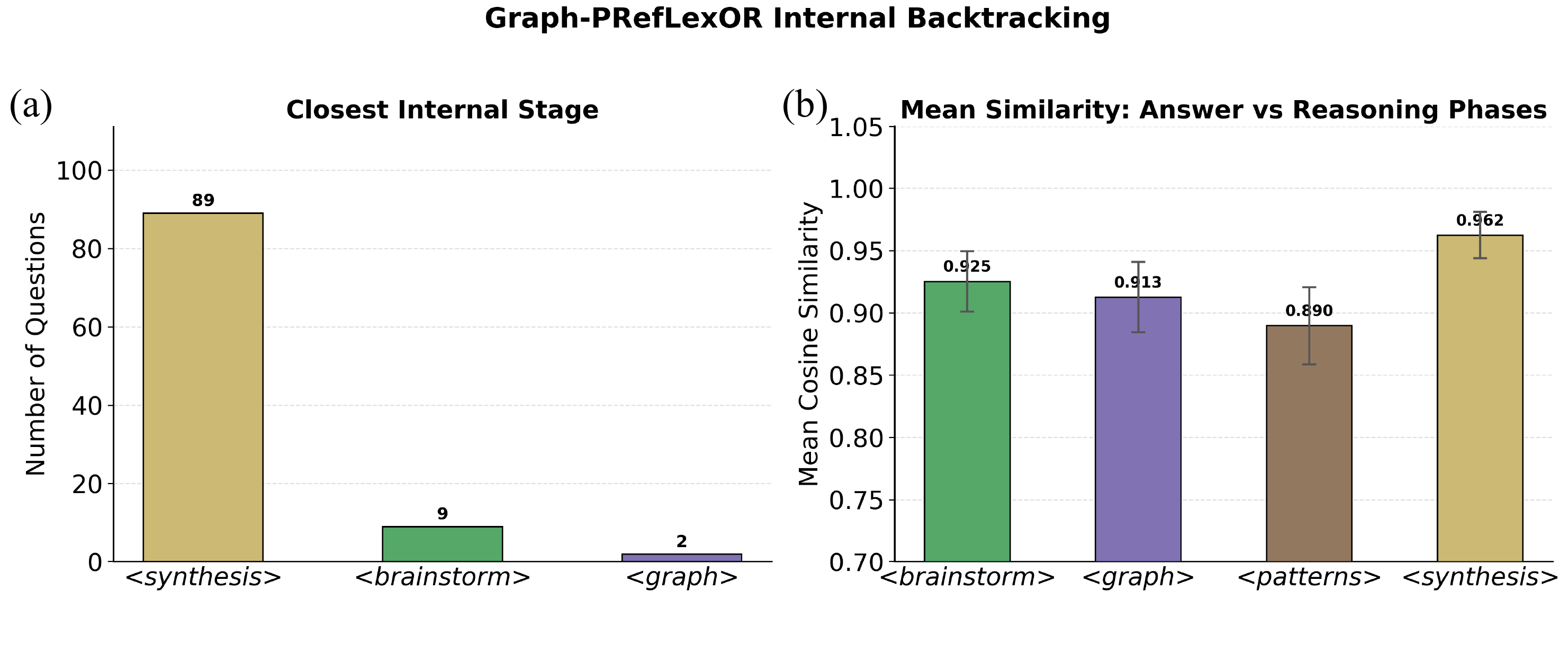}
    \caption{Internal semantic backtracking of Graph-PRefLexOR-8B final answers. (a) Closest structured reasoning stage for each final answer across 100 benchmark questions. (b) Mean cosine similarity between the final answer and each reasoning phase. Final answers align most frequently and most strongly with the \texttt{<synthesis>} stage, indicating that response generation is primarily grounded in the final integrative reasoning step.}
    \label{fig:graph_internal_backtracking}
\end{figure}

As shown in Fig.~\ref{fig:graph_internal_backtracking}a, Graph-PRefLexOR 8B final answers are closest to the \texttt{<synthesis>} stage in 89 of 100 cases, compared with 9 cases for \texttt{<brainstorm>} and 2 cases for \texttt{<graph>}. The mean similarity analysis shows the same trend, with \texttt{<synthesis>} exhibiting the highest average cosine similarity to the final answer, followed by \texttt{<brainstorm>}, \texttt{<graph>}, and \texttt{<patterns>}. This is consistent with the intended role of \texttt{<synthesis>} as the final integrative stage, where candidate mechanisms, graph-encoded relations, and extracted patterns are consolidated into answer-ready prose. Together, these results indicate that Graph-PRefLexOR-8B provides stronger reasoning-answer alignment than Qwen3-8B. While Qwen3-8B often reaches final answers that are semantically close to Graph-PRefLexOR 8B outputs, its own thinking trace is rarely the closest semantic source. By contrast, Graph-PRefLexOR-8B final answers are consistently grounded in its structured reasoning pathway, especially the \texttt{<synthesis>} stage. This supports the conclusion that graph-native reasoning improves not only the quality of intermediate reasoning traces, but also the coherence between reasoning and final response generation.

\subsection{Layer-Wise Reasoning-Answer Divergence} \label{sec:layerwise_divergence} 

We next examine whether the semantic alignment observed at the embedding level is reflected in the models' internal hidden-state representations. Layer-wise hidden-state analysis is commonly used to study how transformer representations evolve across depth~\cite{2020arXiv200212327R,2019arXiv190604341C}, while representation-similarity methods provide a general framework for comparing neural states across layers and conditions~\cite{2017arXiv170605806R,2019arXiv190500414K}. This analysis is particularly relevant here because visible reasoning traces may not always faithfully support the final answer~\cite{lanham_measuring_2023,2024arXiv240213950P,2024arXiv240610625H}. For Qwen3-8B, we compute the cosine distance between hidden states averaged over thinking tokens and hidden states averaged over final-answer tokens at each transformer layer. For Graph-PRefLexOR-8B, we compute the analogous distance between hidden states from the structured reasoning trace and those from the final answer. This yields a layer-wise measure of how strongly each model separates intermediate reasoning from final response generation.

\begin{figure}[!ht]
    \centering
    \includegraphics[width=0.7\linewidth]{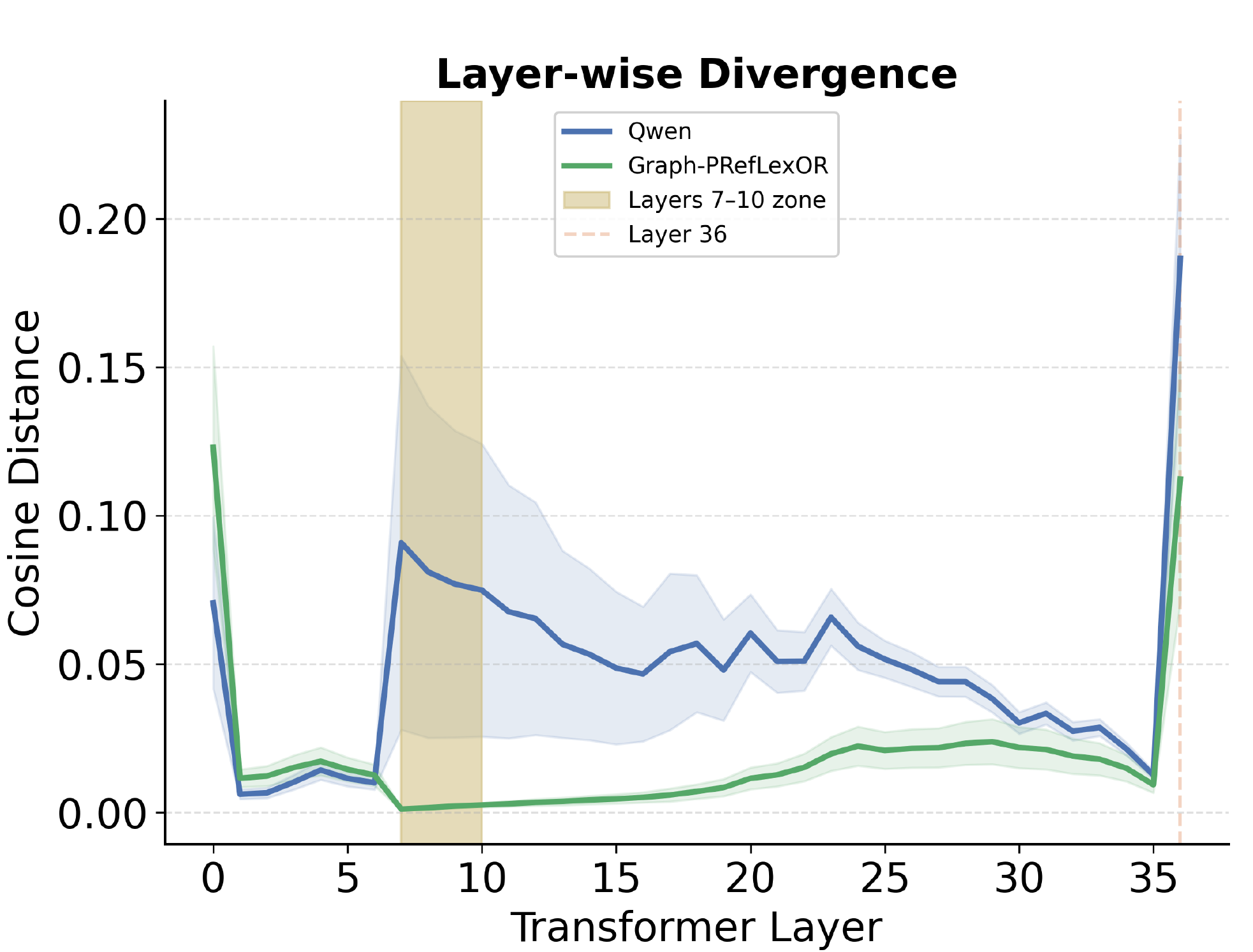}
    \caption{Layer-wise hidden-state divergence between reasoning and final-answer representations for Qwen3-8B and Graph-PRefLexOR-8B. Qwen3-8B exhibits a larger reasoning-answer separation, with a pronounced increase around layers 7-10 and a final-layer spike. In contrast, Graph-PRefLexOR-8B maintains lower divergence across most layers, indicating a more continuous transition from structured reasoning to final-answer generation. Shaded regions denote $\pm$ one standard deviation across questions.}
    \label{fig:qwen_graph_layer_divergence}
\end{figure}

Figure~\ref{fig:qwen_graph_layer_divergence} shows that Qwen3-8B exhibits a substantially larger gap between thinking-state and answer-state representations than Graph-PRefLexOR-8B. The divergence increases sharply around layers 7-10 and rises again at the final layer, suggesting that the baseline model undergoes a stronger representational shift between visible reasoning and final answer generation. By contrast, Graph-PRefLexOR-8B maintains a smaller reasoning-answer distance across most layers, consistent with a more continuous transition from structured intermediate reasoning to the final response. To connect this layer-wise behavior with the semantic backtracking results, we further quantify examples according to whether the final answer backtracks to the model's own reasoning. For Qwen3-8B, we divide the benchmark into cases where the final answer is closest to its own \texttt{<think>} trace and cases where it is closest to another reference source. For Graph-PRefLexOR-8B, we analogously separate cases where the final answer aligns with its own structured reasoning stages from cases where it aligns with Qwen-derived outputs.

\begin{figure}[!ht]
    \centering
    \includegraphics[width=\linewidth]{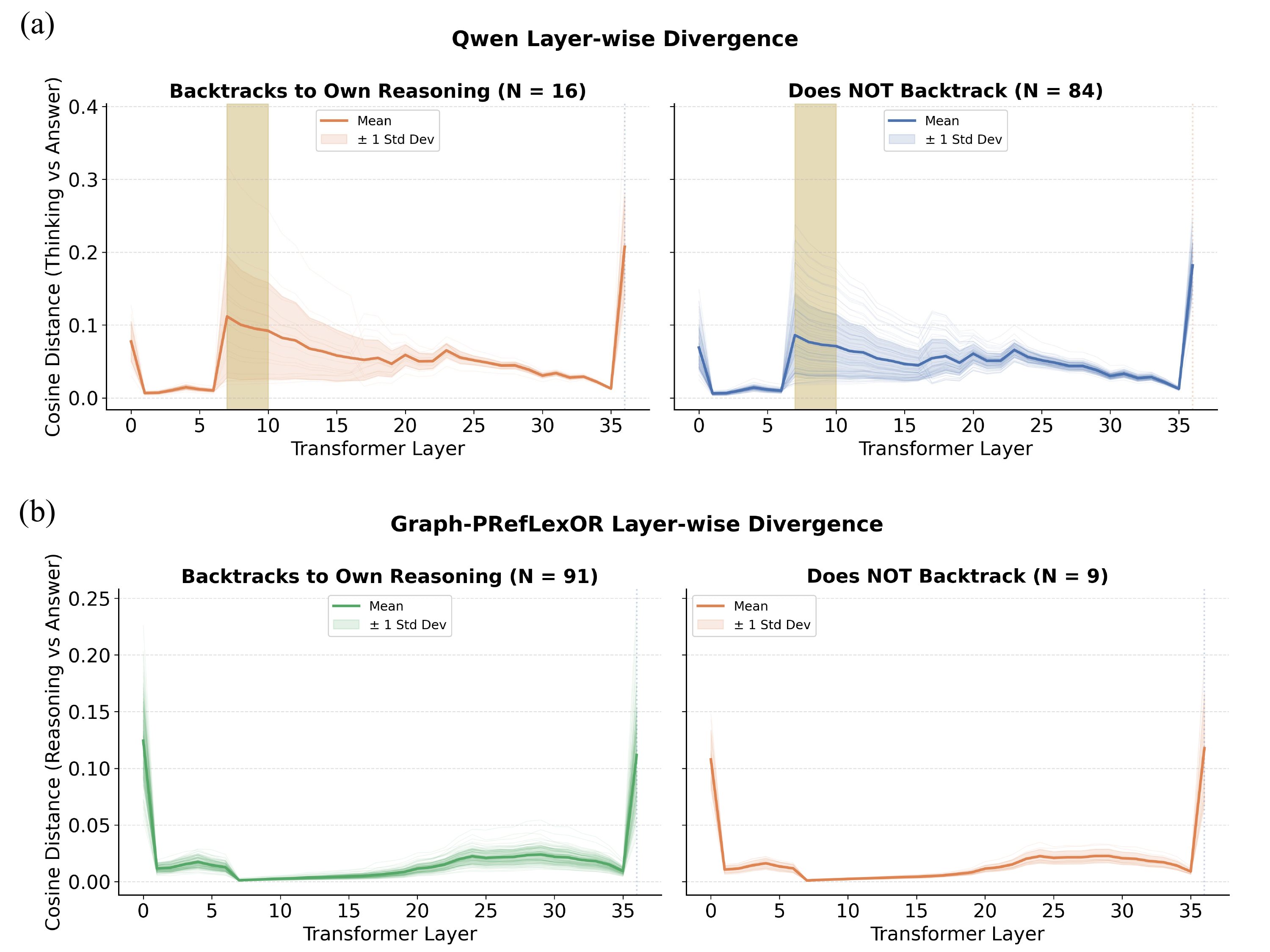}
    \caption{Backtracking-conditioned layer-wise hidden-state divergence for Qwen3-8B and Graph-PRefLexOR-8B. (a) Qwen3-8B divergence between thinking and final-answer, separated by whether the final answer backtracks to the model's own thinking trace or to another source. Non-backtracking cases show larger divergence, particularly around layers 7-10 and at the final layer. (b) Graph-PRefLexOR-8B divergence between structured reasoning and final-answer states, separated by whether the final answer backtracks to its own reasoning stages or to another source. Graph-PRefLexOR maintains lower divergence across both groups, indicating stronger continuity between structured reasoning and final-answer generation. Shaded regions denote $\pm$ one standard deviation across questions.}
    \label{fig:graph_stage_divergence}
\end{figure}

Figure~\ref{fig:graph_stage_divergence}a shows that the Qwen3-8B reasoning-answer gap depends strongly on backtracking behavior. When the final answer backtracks to Qwen's own thinking trace, hidden-state divergence is lower. When it does not, the divergence is larger, with the clearest separation again emerging around layers 7-10 and at the final layer. This suggests that layers 7-10 mark an early transition region where the final-answer pathway begins to separate from the visible thinking trace. We discuss in detail about this in the Supplementary Information. Graph-PRefLexOR 8B shows a more stable pattern across backtracking groups (See Figure \ref{fig:graph_stage_divergence}b). Most examples backtrack to the model's own structured reasoning stages, and both groups maintain lower divergence than Qwen3-8B. This indicates that Graph-PRefLexOR-8B preserves a tighter representational connection between intermediate reasoning and final-answer generation, even when examples are stratified by alignment source. Together, these hidden-state analyses support the semantic backtracking results. Qwen3-8B can produce final answers that are semantically close to Graph-PRefLexOR-8B outputs, but its internal representations often show a larger separation between visible thinking and final response generation. In contrast, Graph-PRefLexOR-8B maintains lower layer-wise divergence and stronger continuity between structured reasoning and final-answer states. This suggests that graph-native reasoning improves not only the interpretability of the generated trace, but also the internal stability of the reasoning-to-answer pathway.

\subsection{Test-Time Graph Expansion}

\begin{figure*}[!ht]
  \centering
  \includegraphics[width=.8\textwidth]{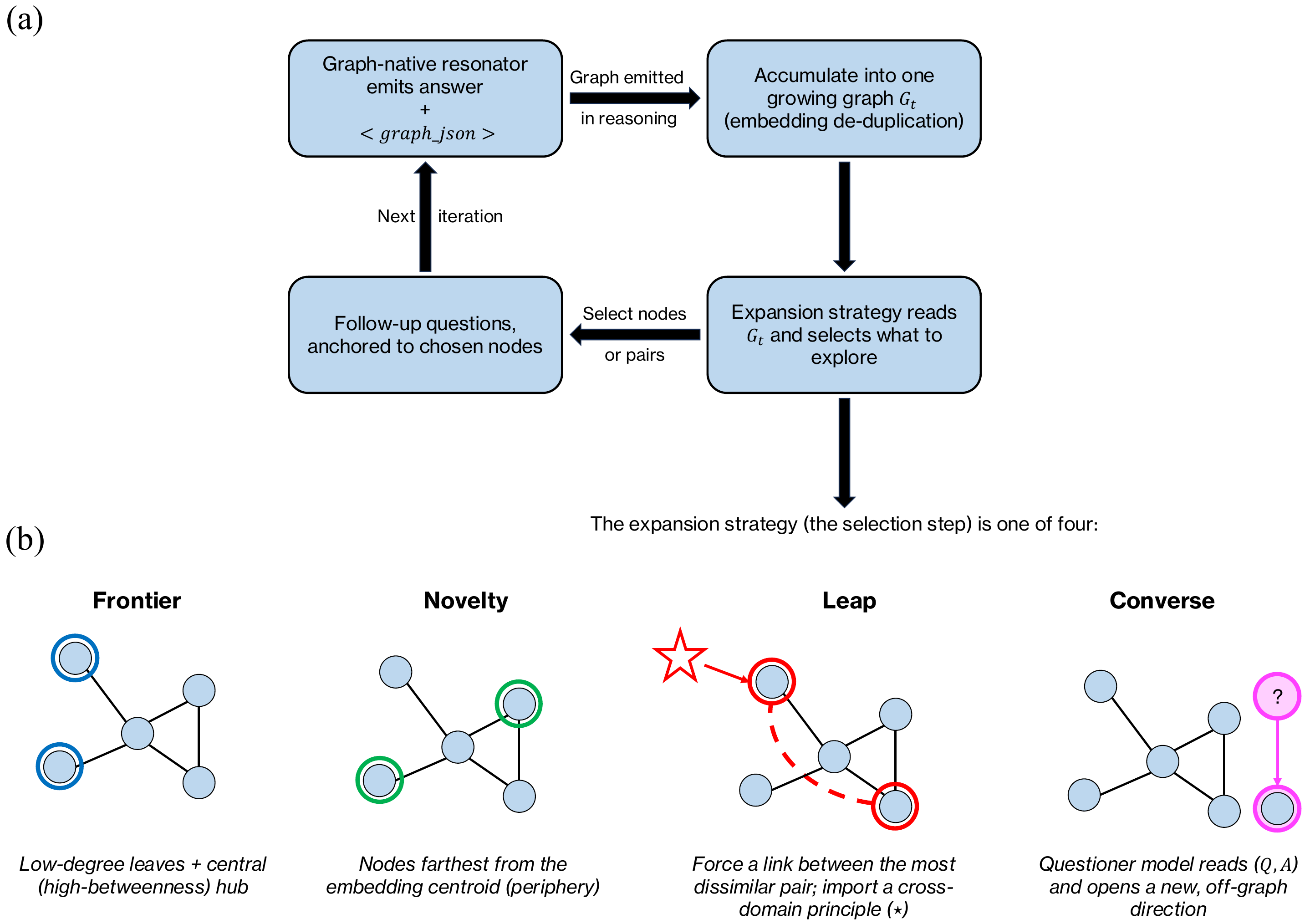}
  \caption{Graph-native ideation loop for test-time graph expansion. At each iteration, the reasoner answers a question, emits a small ontological graph, and merges it into a growing memory graph $G_t$ using embedding-based de-duplication. An expansion strategy then selects concepts or concept pairs from $G_t$ to generate the next question. The four strategies are \emph{frontier}, which expands low-degree leaves and central hubs; \emph{novelty}, which targets embedding-peripheral concepts; \emph{leap}, which forces distant recombination and cross-domain import; and \emph{converse}, which uses a questioner model to introduce off-graph directions.}
  \label{fig:approach}
\end{figure*}

The preceding sections show that Graph-PRefLexOR produces structured reasoning traces, broader semantic exploration, and stronger reasoning--answer alignment in single-response settings. We next ask whether this graph-native reasoning format can also support iterative scientific ideation when additional test-time compute is available. To test this, we convert the reasoner into a self-expanding graph engine (Fig.~\ref{fig:approach}). At each iteration, the model answers a question and emits a small ontological graph within its reasoning trace. This graph is merged into a growing memory graph, $G_t$, using embedding-based de-duplication, allowing accumulated graph structure rather than the context window alone to carry information across iterations. An expansion strategy then reads $G_t$ and selects concepts or concept pairs to guide the next question. We compare four strategies: \emph{frontier} expands low-degree leaves and the central high-betweenness hub; \emph{novelty} targets the nodes farthest from the embedding centroid; \emph{leap} forces a mechanism between the most dissimilar concepts and imports a principle from an unrelated field ($\star$); and \emph{converse} uses a separate questioner model to introduce a new concept the graph does not yet contain. Iterating this loop tests whether additional inference-time computation expands, densifies, or recombines the model's scientific idea space. Further detail about these strategies are given in the Methods section.

\begin{figure*}[!ht]
  \centering
  \includegraphics[width=.9\textwidth]{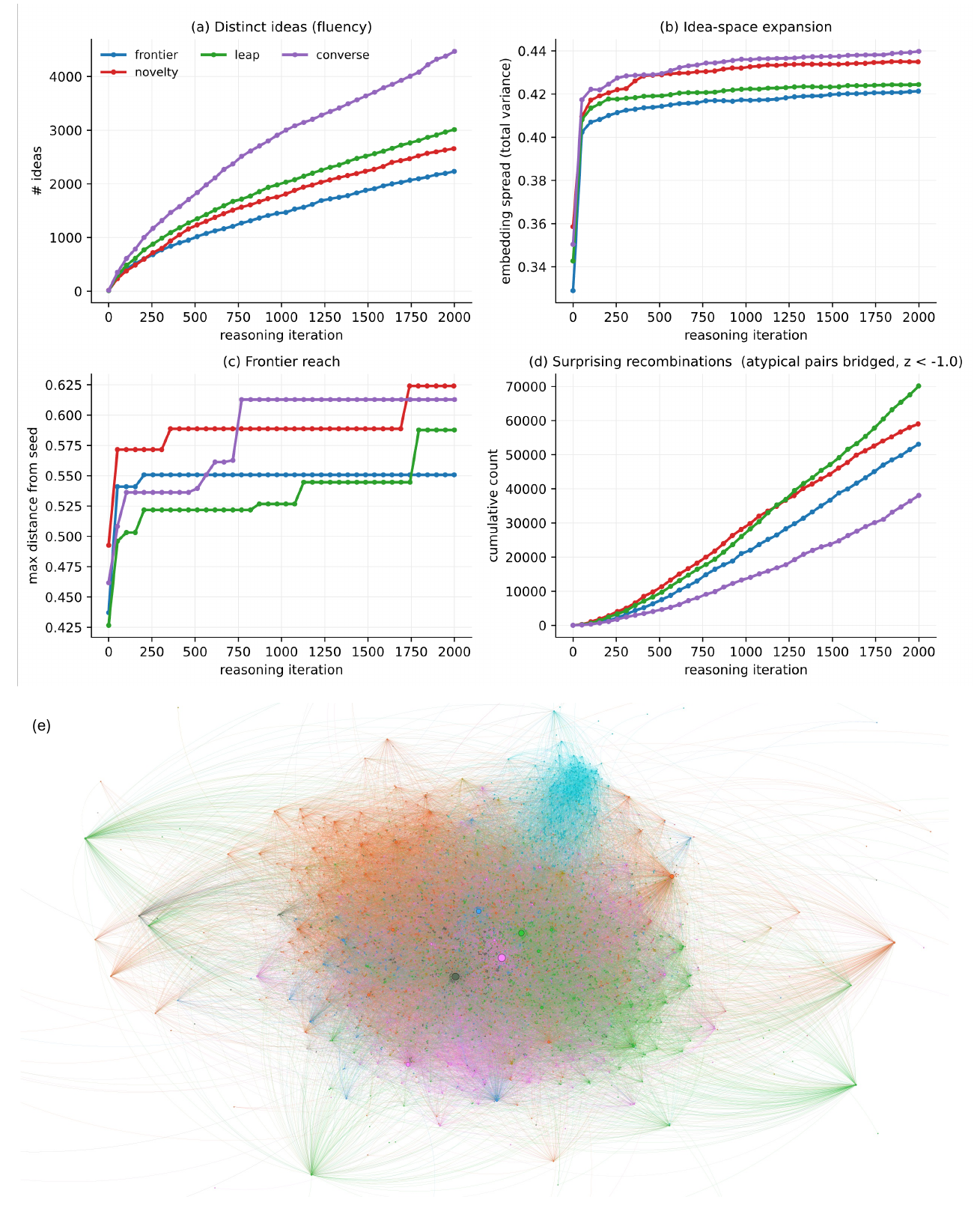}
  \caption{Test-time compute expands a bounded idea space through recombination. Four size-robust metrics are shown as a function of reasoning iteration up to $2{,}000$ iterations for the four expansion strategies. The number of distinct concepts continues to increase (a), whereas the explored embedding volume (b) and maximum distance from the seed (c) saturate within a few hundred iterations, indicating that the semantic territory of a fixed topic is bounded and rapidly covered. In contrast, surprising recombinations (d), defined as atypically dissimilar concept pairs ($z<-1$ relative to the global pairwise-similarity null) later bridged through a shared intermediate concept, grow super-linearly and most rapidly under \emph{leap}. Panel (e) shows the resulting \emph{leap} graph after $3{,}700$ iterations, with node color denoting modularity class and node size denoting degree ($4{,}419$ nodes and $37{,}064$ edges). All reported $z$ values denote standardized deviations from the corresponding randomized null distribution.}
  \label{fig:900}
\end{figure*}

\begin{figure*}[!ht]
  \centering
  \includegraphics[width=.8\textwidth]{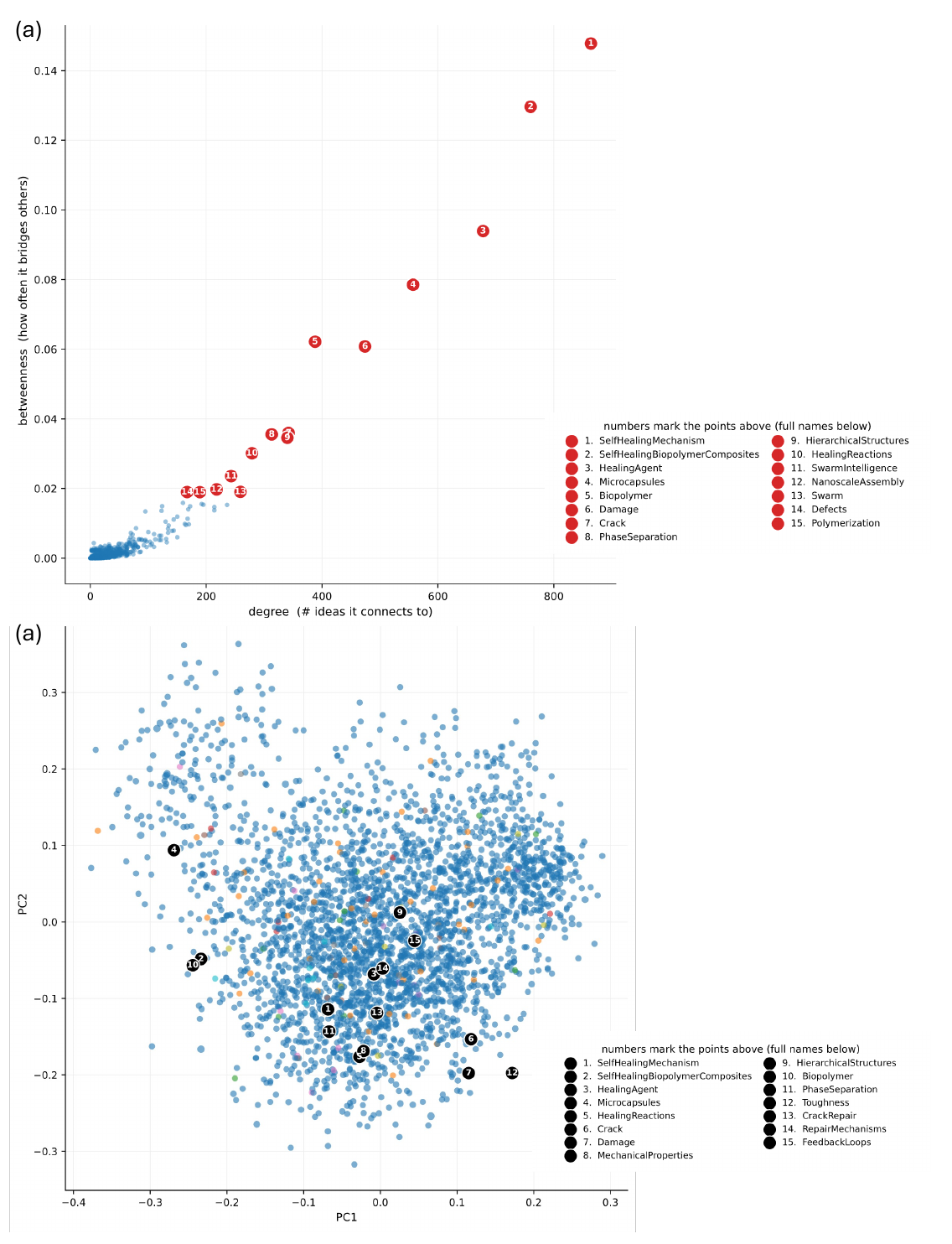}
  \caption{Semantic organization and broker concepts in the \emph{leap} run. (a) Principal-component projection of concept embeddings, with colors indicating greedy-modularity communities and marker size indicating PageRank. The fifteen highest-PageRank concepts are numbered and listed below the map. (b) Broker concepts plotted by degree and betweenness. A small set of high-betweenness concepts mediates most cross-community bridging, including cross-domain imports such as swarm intelligence and nanoscale assembly.}
  \label{fig:900-part-2}
\end{figure*}

Treating the model as a self-expanding ontological representation of thinking reveals that additional test-time compute produces recombinational rather than purely expansive growth (Fig.~\ref{fig:900}). Across all expansion strategies, the number of distinct concepts continues to increase, but both the explored embedding volume and the maximum distance from the seed plateau within a few hundred iterations. This indicates that the semantic territory associated with a fixed topic is bounded and rapidly covered. In contrast, the cumulative number of surprising recombinations, defined as atypically dissimilar concept pairs that are later bridged through a shared intermediate, continues to grow super-linearly through $2{,}000$ iterations. Thus, additional compute does not primarily discover ever more distant regions of semantic space; instead, it densifies a bounded idea space by creating new bridges among already accessible concepts. The expansion strategies differ most clearly along this recombinational axis. The divergent \emph{leap} policy converts compute into surprising recombinations most efficiently, whereas \emph{converse} produces the largest number of concepts but the fewest bridges, indicating that concept generation and recombinational synthesis are distinct capabilities. The resulting idea space is organized around a small set of high-centrality hubs (Fig.~\ref{fig:900-part-2}a), while high-betweenness broker concepts, including cross-domain imports such as swarm intelligence and nanoscale assembly, mediate much of the bridging between sub-fields (Fig.~\ref{fig:900-part-2}b).

\begin{figure*}[ht]
  \centering
  \includegraphics[width=.9\textwidth]{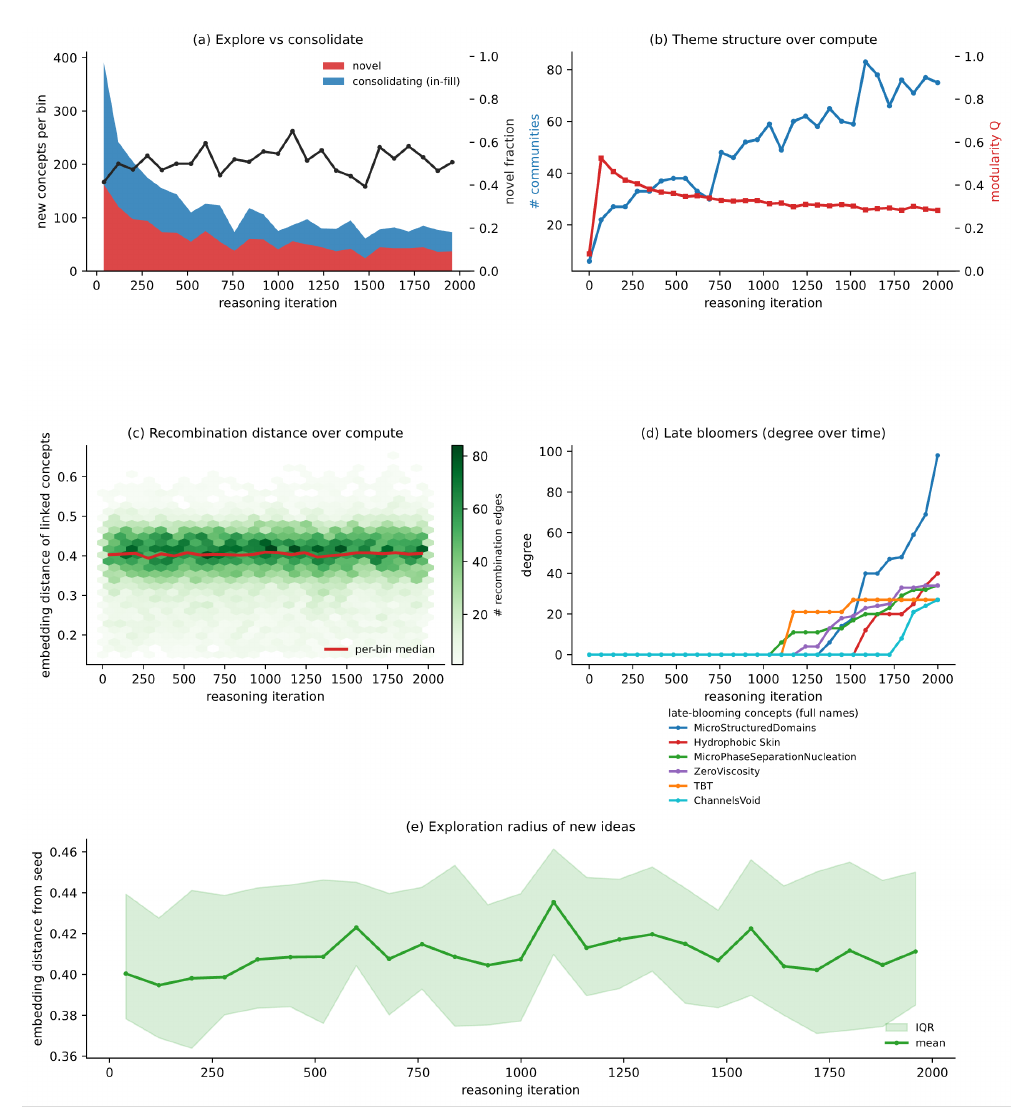}
  \caption{Growth dynamics of the \emph{leap} run. The final graph is replayed in birth-iteration order, and each panel is evaluated using embedding geometry or mesoscale community structure rather than raw graph distance. (a) New concepts per iteration bin, separated into novel concepts and consolidating in-fill concepts; the black line shows the fraction of novel concepts. (b) Number of greedy-modularity communities and modularity $Q$. (c) Embedding distance between the endpoints of each recombination edge as a function of iteration. (d) Degree trajectories of six late-blooming concepts that acquire most of their final degree in the second half of the run. (e) Mean embedding distance of newly added concepts from the seed, with interquartile range. Together, the panels show sustained novelty, stable semantic reach, increasing cross-community interconnection, and delayed activation of high-impact concepts.}
  \label{fig:901}
\end{figure*}

The dynamics of the \emph{leap} run (Fig.~\ref{fig:901}) indicate a steady exploratory regime rather than a transition from exploration to consolidation. Although the rate of concept addition decreases as the topic saturates, the fraction of genuinely novel concepts remains close to one half throughout (Fig.~\ref{fig:901}a). At the same time, newly introduced concepts continue to appear at an approximately constant embedding distance from the seed (Fig.~\ref{fig:901}e), and the semantic span of recombination edges remains similarly stable (Fig.~\ref{fig:901}c). Thus, \emph{leap} does not converge inward; instead, it persistently recombines concepts across a bounded semantic radius.

The structural signature of this process is mesoscale densification. The number of communities increases from roughly six to seventy-five, while modularity declines steadily (Fig.~\ref{fig:901}b), indicating that sub-fields continue to proliferate even as they become increasingly interconnected. Growth is also cumulative rather than strictly sequential: several concepts introduced early remain dormant for more than $1{,}000$ iterations before abruptly emerging as hubs (Fig.~\ref{fig:901}d). This shows that the system repeatedly revisits and amplifies earlier ideas. Together, these mechanisms explain the scaling behavior in Fig.~\ref{fig:900}: within a bounded semantic territory that is continuously and evenly recombined, the number of bridged concept pairs, and hence the count of surprising recombinations, continues to increase without clear saturation.

\begin{figure}[t]
  \centering
  \includegraphics[width=.9\linewidth]{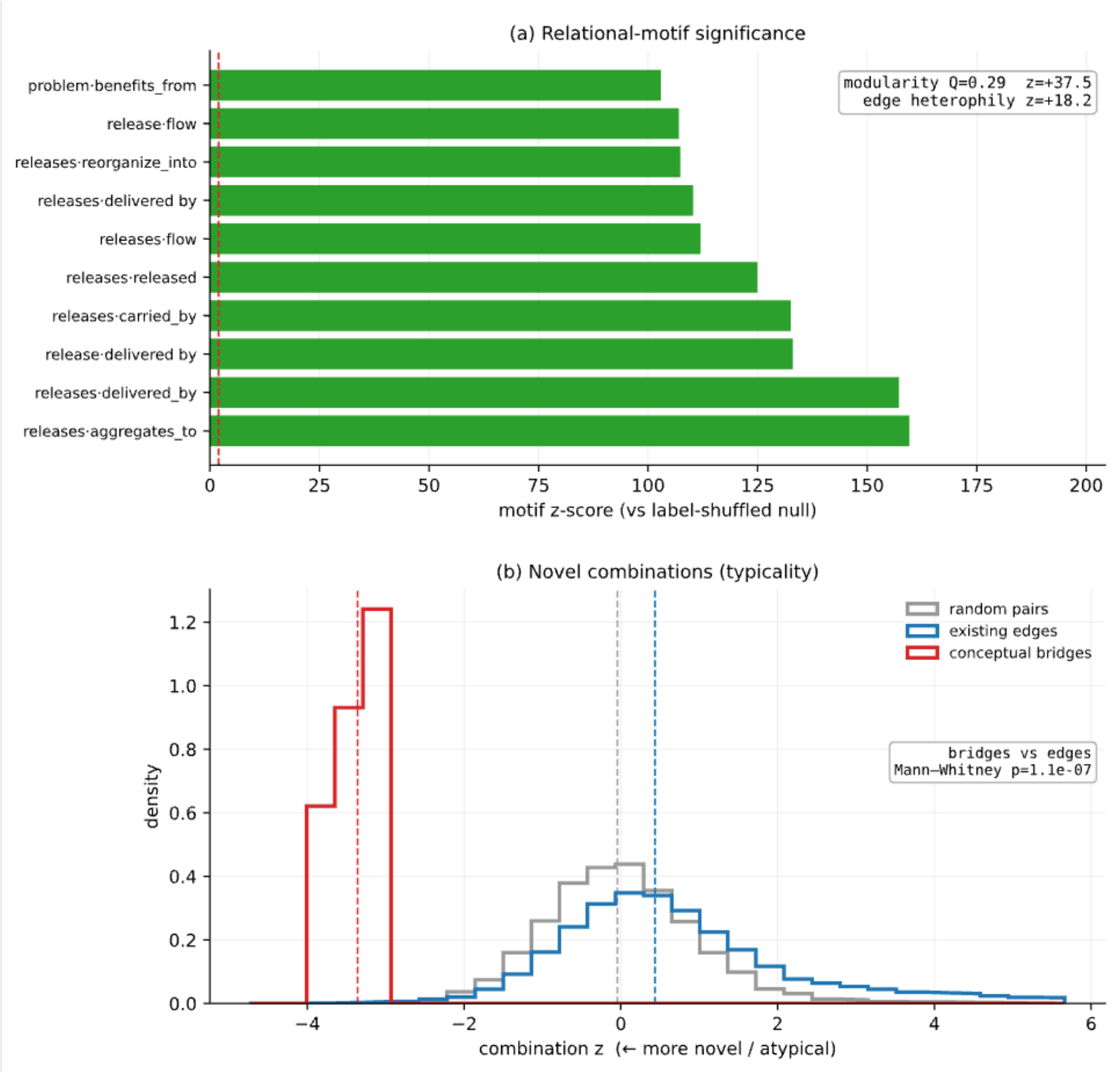}
  \caption{Statistical novelty of mined connections in the \emph{leap} run. (a) Relational-motif significance relative to a label-shuffled null model. The ten most over-represented relation-typed two-step motifs reach $z\approx100$--$160$, far exceeding the ordinary significance threshold ($z=1.96$), indicating that the graph follows consistent relational templates rather than random associations. The graph is also more modular than chance ($Q=0.29$, $z=+37.5$) and preferentially connects semantically dissimilar concepts (edge heterophily $z=+18.2$). (b) Combination typicality analysis comparing random concept pairs, direct graph edges, and two-hop conceptual bridges. Direct edges are mildly typical (median $z\approx+0.4$), whereas two-hop bridges lie in the atypical tail (median $z\approx-3.4$; Mann--Whitney $p=1.1\times10^{-7}$), showing that novelty is concentrated in long-range conceptual recombinations rather than direct links.}
  \label{fig:902}
\end{figure}

The connections formed by the model are statistically novel rather than artifacts of graph size (Fig.~\ref{fig:902}). To test this, we compare the accumulated graph against randomized null models that preserve its size and wiring. First, recurring relational motifs, defined as relation-typed two-step patterns, are strongly over-represented relative to a label-shuffled null model ($z\approx100$--$160$). This indicates that the graph reuses structured relational templates rather than forming arbitrary associations. The graph also exhibits substantially higher modularity than expected by chance ($Q=0.29$, $z=+37.5$), while its edges preferentially connect semantically dissimilar concepts (edge heterophily $z=+18.2$). Thus, the accumulated structure is both organized and heterophilic, combining thematic modularity with cross-concept linkage.

The location of novelty is more specific. We quantify each concept pairing using combination typicality, which measures how unusual a pair of concepts is relative to the global concept-similarity distribution; more negative values indicate more atypical combinations. Direct edges are mildly homophilic and slightly more typical than random concept pairs (median $z\approx+0.4$), suggesting that explicit graph links often connect related ideas. In contrast, conceptual bridges across graph distance two lie deep in the atypical tail (median $z\approx-3.4$; Mann--Whitney $p=1.1\times10^{-7}$), indicating that they connect concepts that are normally unrelated. Novelty therefore resides primarily in the long-range recombinations implied by the graph, rather than in the direct edges themselves. These long-range bridges are the same structures whose cumulative count continues to grow with additional test-time compute (Fig.~\ref{fig:900}).

\section{Conclusion}
In this work, we introduced a new generation of Graph-PRefLexOR models trained with Group Relative Policy Optimization (GRPO) for traceable scientific hypothesis generation in materials design. Building on earlier Graph-PRefLexOR formulations that mapped a task to a knowledge graph, abstract patterns, and a final answer, the present models use a more granular sentinel-based reasoning format consisting of \texttt{<brainstorm>}, \texttt{<graph>}, \texttt{<graph\_json>}, \texttt{<patterns>}, and \texttt{<synthesis>} stages. This structure decomposes open-ended scientific reasoning into mechanism exploration, concept abstraction, machine-readable graph construction, pattern extraction, and final hypothesis synthesis. Unlike earlier ORPO/EXO-style preference-optimization approaches, the present models use GRPO to optimize structured reasoning behavior, providing an interpretable bridge between neural language generation and symbolic knowledge representation.

Across a manually curated benchmark of 100 open-ended questions derived from materials science and mechanics literature, Graph-PRefLexOR consistently outperformed its corresponding base models in reasoning quality, intellectual depth, and reasoning traceability. The strongest improvements were observed in traceability, indicating that graph-structured reasoning primarily enhances the organization and causal transparency of intermediate reasoning. Comparisons with no-thinking baselines further show that these gains arise from explicit intermediate reasoning rather than model scale or architecture alone. Embedding-based analyses support this conclusion: Graph-PRefLexOR reasoning traces occupy broader and more structured semantic regions, follow more coherent stage-wise trajectories, and exhibit approximately $2\times$--$3\times$ greater semantic diversity than base-model traces.

The reasoning--answer alignment analyses show that these structured traces are not merely decorative. Semantic backtracking indicates that Qwen3-8B final answers align with their own visible thinking traces in only a minority of cases and more often occupy semantic regions closer to Graph-PRefLexOR-derived outputs. In contrast, Graph-PRefLexOR final answers remain strongly anchored to their own structured reasoning pathway, most frequently backtracking to the \texttt{<synthesis>} stage. Layer-wise hidden-state analyses reinforce this pattern: Qwen3-8B exhibits larger reasoning--answer divergence, particularly around layers 7--10 and at the final layer, whereas Graph-PRefLexOR maintains lower divergence across layers and backtracking groups. Together, these results suggest that graph-native reasoning improves not only the visible organization of reasoning traces, but also the semantic and representational continuity between reasoning and final answer generation.

We further showed that graph-native reasoning can be extended from single-response hypothesis generation to iterative test-time ideation. By accumulating the model's emitted \texttt{<graph\_json>} outputs into a growing graph memory and using expansion policies to select follow-up questions, Graph-PRefLexOR becomes a self-expanding graph engine. This analysis reveals that additional test-time compute does not simply expand semantic territory indefinitely. Instead, the explored embedding volume and maximum distance from the seed saturate rapidly, while the number of surprising recombinations continues to grow super-linearly. The most divergent expansion strategy, \emph{leap}, is especially effective at converting compute into long-range conceptual bridges, showing that concept generation and recombinational synthesis are distinct capacities.

Statistical null-model analyses confirm that the resulting idea graphs are not artifacts of graph size. Relation-typed motifs are strongly enriched relative to randomized label assignments, the graph exhibits modular organization beyond chance, and edge heterophily indicates systematic linking of semantically dissimilar concepts. Most importantly, novelty is concentrated not in direct graph edges, which remain mildly homophilic, but in two-hop conceptual bridges that connect normally unrelated concepts. This finding reframes scientific ideation as the progressive densification of a bounded semantic space: test-time compute increases the number of meaningful bridges among accessible concepts, rather than merely pushing the model toward ever more distant regions.

Taken together, these results establish Graph-PRefLexOR as a compact and interpretable framework for scientific hypothesis generation. The model improves final reasoning performance by restructuring the intermediate pathway through which answers are produced, grounding final responses in graph-native reasoning stages, and enabling iterative recombination through accumulated graph memory. More broadly, this work emphasizes that scientific language models should be evaluated not only by final-answer quality, but also by reasoning traceability, semantic diversity, reasoning--answer alignment, representation dynamics, and test-time recombinational growth. Future work will extend Graph-PRefLexOR toward autonomous discovery workflows that integrate literature reasoning, simulation, experimental feedback, and closed-loop design of polymers, composites, and multifunctional materials across scales.

\section{Materials and Methods}

\subsection{Training Strategy}

\label{sec:training-graphpreflexor}

We train a family of graph-native reasoning models, Graph-PRefLexOR at $1.7$B, $3$B, and
$8$B parameters (Table~\ref{tab:model_hf_ids}), with a single two-stage recipe applied to
three instruction-tuned backbones that differ in an important respect. The Qwen3-1.7B and
Qwen3-8B backbones are already reasoning models with a native thinking mode, so the
recipe adapts an existing reasoning ability to our specific graph-native format. The
Llama-3.2-3B-Instruct backbone, by contrast, is a standard instruction-tuned model
with no built-in reasoning behavior; the 3B therefore represents the harder case of
converting an ordinary chat model into a graph-native reasoner essentially from scratch. As we show in \S\ref{subsec:gp-results}, this distinction (not merely parameter count) is what most strongly governs each model's training dynamics. We first describe how the training corpus is constructed (\S\ref{subsec:gp-data}), then the two-stage training approach (\S\ref{subsec:gp-approach}), the composite reward (\S\ref{subsec:gp-reward}), and the resulting training dynamics (\S\ref{subsec:gp-results}).

\subsubsection{Dataset Construction}
\label{subsec:gp-data}

The training corpus is built by a teacher-distillation pipeline that
converts raw text into structured, graph-native preference pairs. Source passages are
sampled from a mixture of streaming corpora (general educational web text
(\texttt{fineweb-edu}) and a domain-specific biological and mechanical-materials mixture
(\texttt{bio-silk-mech-mix-80K})) giving both breadth and a materials-science focus. For
each passage, a strong teacher model (GPT-5.1) first writes a single challenging,
self-contained, expert-level question grounded in the passage, and then produces the
preferred (chosen) response: a complete graph-native reasoning trace that fills the exact sentinel template of \S\ref{subsec:gp-approach}: \texttt{<brainstorm>}
$\rightarrow$ \texttt{<graph>} $\rightarrow$ \texttt{<graph\_json>} (a strict
\texttt{nodes}/\texttt{edges} JSON) $\rightarrow$ \texttt{<patterns>} $\rightarrow$
\texttt{<synthesis>}, closed by \texttt{</think>} and followed by a thorough final
answer, under a system prompt instructing it to ``reason using a graph-based latent
structure.'' A separate, deliberately weak model (GPT-5-nano), prompted to give a hurried
$1$--$3$ sentence answer with no reasoning and no special tokens, produces the
dispreferred (rejected) response. The gold answer is the text following
\texttt{</think>}, and any example whose \texttt{graph\_json} fails to parse or whose answer
is empty is discarded, so every retained record carries a valid graph and a complete answer.
Each record stores the question (prompt), the gold answer, the chosen and rejected
completions, and the extracted teacher graph.

This construction directly shapes both training stages. ORPO (\S\ref{subsec:gp-approach})
consumes the chosen/rejected pairs, so by design it contrasts a rich, structured
graph-reasoning trace against a shallow direct answer (a deliberately large behavioral gap
that, as we show in \S\ref{subsec:gp-results}, makes the preference ordering easy to learn
and the cold-start preference accuracy saturate quickly. Graph-GRPO consumes only the prompt
and gold answer (the chosen/rejected fields are unused), letting the policy explore its own
traces while the composite reward) built on the same \texttt{graph\_json} object that the
teacher template seeds grades correctness and graph quality. Distilling from a strong
teacher means the target behavior is demonstrated rather than discovered, which is
what lets even the small backbones acquire the structured format in a single epoch.

\subsubsection{Training Approach}
\label{subsec:gp-approach}

All sizes are trained with the same two-stage recipe
(Table~\ref{tab:gp-hparams}). The model emits a structured, graph-native reasoning trace
rather than free-form chain-of-thought, following the recursive, reflective paradigm of
PRefLexOR~\cite{buehler_preflexor_2025}: all deliberation is enclosed in a
\texttt{<think>}\dots\texttt{</think>} block containing, in order, a \texttt{<brainstorm>}
(open exploration), a \texttt{<graph>} (a natural-language sketch of entities and relations),
a \texttt{<graph\_json>} (a machine-readable knowledge graph with typed nodes and
\texttt{source}/\texttt{relation}/\texttt{target} edges), a \texttt{<patterns>} section
(motifs read off the graph), and a \texttt{<synthesis>}; the final answer is emitted
after the closing \texttt{</think>} tag.
All runs used a single GPU; the 1.7B and 8B (Qwen3) models were trained on an NVIDIA A100-SXM4-80GB, while the 3B (Llama-3.2-3B-Instruct) model was trained on an NVIDIA GH200 480GB.

\begin{figure*}[t]
    \centering
    \includegraphics[width=\textwidth]{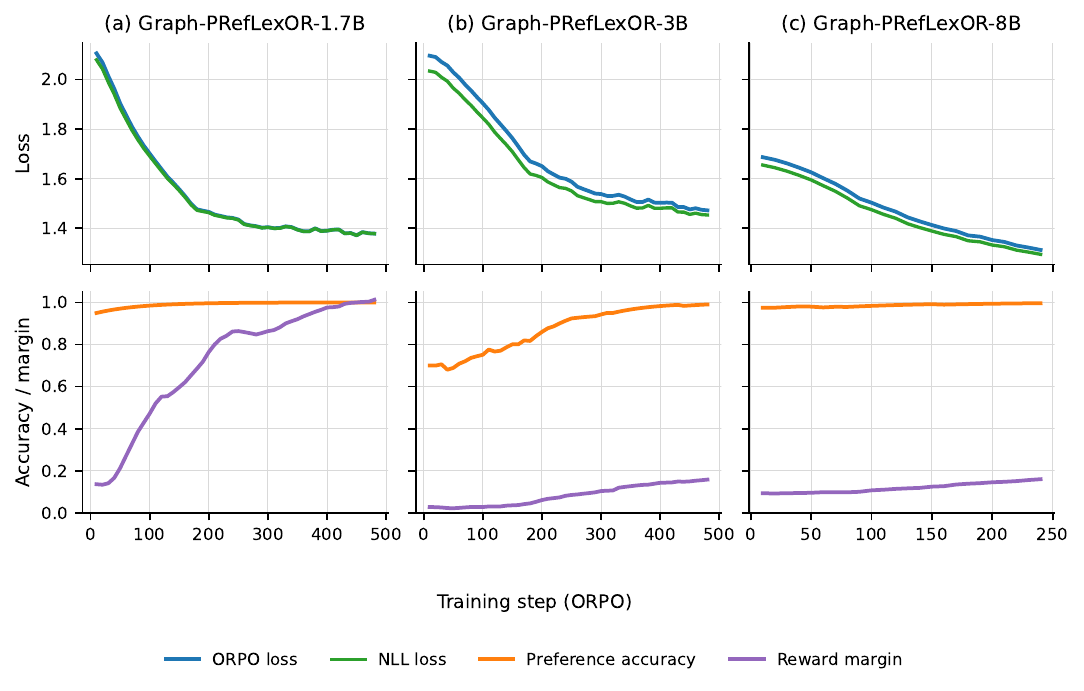}
    \caption{ORPO cold start for all three models. Top row: total ORPO loss and its
    NLL component; bottom row: preference accuracy and reward margin (both in
    $[0,1]$); columns are \textbf{(a)} 1.7B, \textbf{(b)} 3B, \textbf{(c)} 8B, with $y$ shared
    per row (note the differing ORPO durations, $\sim\!480$/$480$/$240$ steps). Loss falls and
    preference accuracy saturates near $1.0$ for all backbones; the 1.7B's much larger reward
    margin reflects its higher learning rate rather than scale.}
    \label{fig:gp-orpo-panels}
\end{figure*}

\begin{figure*}[t]
    \centering
    \includegraphics[width=\textwidth]{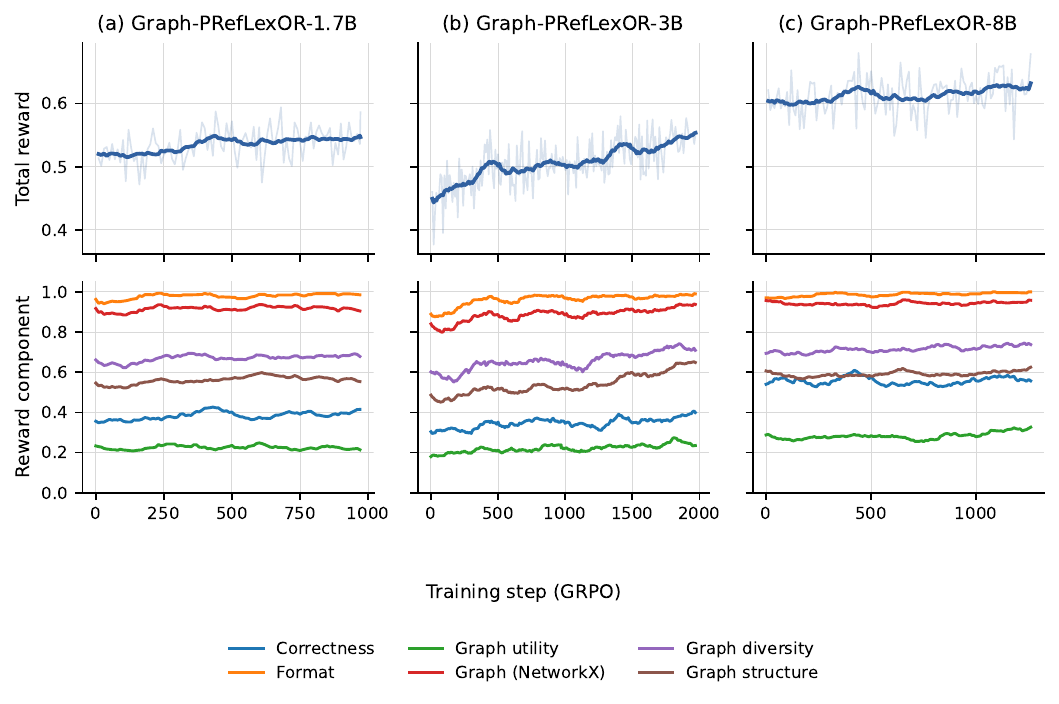}
    \caption{Graph-GRPO reward for all three models. Top row: total composite reward;
    bottom row: the six reward components; columns are \textbf{(a)} 1.7B,
    \textbf{(b)} 3B, \textbf{(c)} 8B (differing GRPO durations, $\sim\!970$/$1970$/$1260$
    steps), with $y$ shared per row for direct comparison. The 8B starts highest and the 3B
    climbs most, while graph utility (green) is the lowest component at every scale.}
    \label{fig:gp-reward-panels}
\end{figure*}

\begin{figure*}[t]
    \centering
    \includegraphics[width=\textwidth]{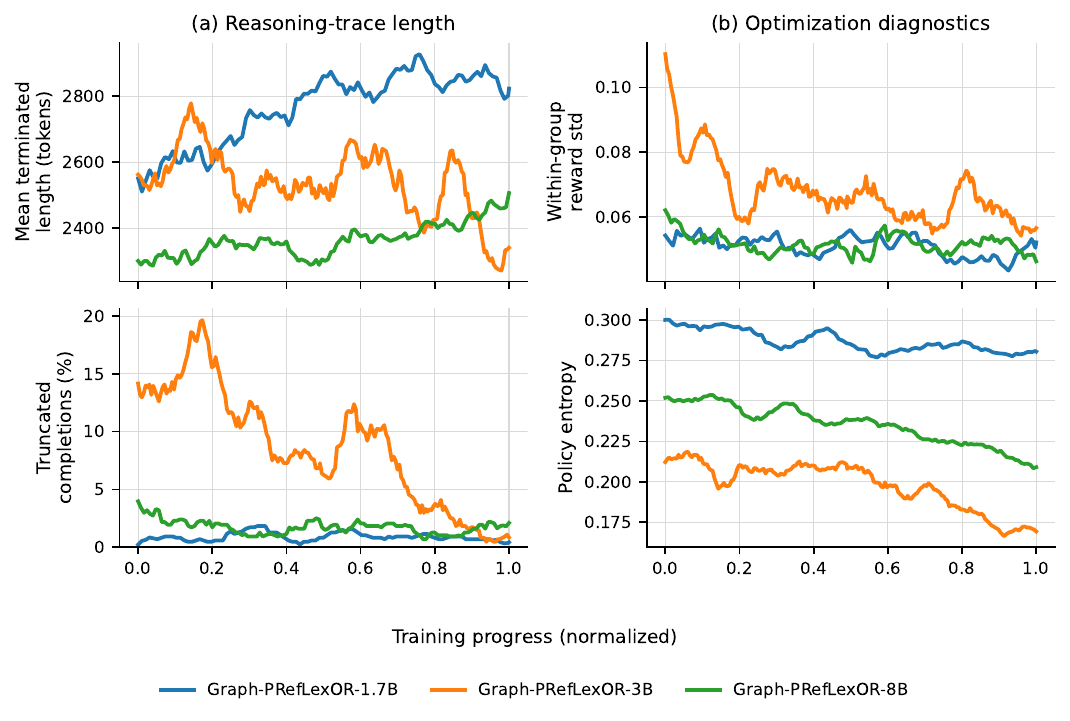}
    \caption{Graph-GRPO dynamics across the three models, with each run rescaled to $[0,1]$
    training progress so the durations align. \textbf{(a)} Reasoning-trace length: mean
    terminated completion length (top) and fraction of completions truncated at the token
    budget (bottom). \textbf{(b)} Optimization diagnostics: within-group reward standard
    deviation, i.e.\ the scale of the group-normalized advantage of Eq.~\eqref{eq:grpo-adv}
    (top), and policy entropy (bottom). We find that the 3B alone learns to fit its budget (truncation
    $20\%\!\rightarrow\!1\%$) and carries the largest reward dispersion.}
    \label{fig:gp-length-diag}
\end{figure*}

\subsubsubsection{Rationale for This Reasoning Structure}
Each of the sentinel blocks are plays a distinct epistemic role, and
their order encodes a deliberate divergent-to-convergent scientific reasoning pipeline: \emph{exploration} $\rightarrow$ \emph{formalization} $\rightarrow$ \emph{abstraction}
$\rightarrow$ \emph{explanation}. \texttt{<brainstorm>} performs a wide search over
hypotheses and candidate mechanisms; \texttt{<graph>} draws a scientifically grounded conceptual
blueprint; \texttt{<graph\_json>} commits it to a canonical, machine-readable knowledge
graph; \texttt{<patterns>} compresses the graph into reusable motifs (causal loops, modular
micro$\rightarrow$meso$\rightarrow$macro hierarchies, bottlenecks, invariants); and
\texttt{<synthesis>} reads an ordered explanation back off the graph. We adopt this
representation for three reasons. First, \emph{relational faithfulness}: scientific reasoning
is intrinsically about entities and their causal and structural relationships, so an explicit
graph is a more natural, inspectable substrate than linear text, and the multi-scale
abstractions targeted by \texttt{<patterns>} are precisely graph-theoretic. Second,
\emph{verifiability and reward access}: by emitting a canonical \texttt{<graph\_json>}, the
reasoning becomes a parseable object that the reward function can interrogate directly---the
validity, structure, diversity, and graph-utility terms of \S\ref{subsec:gp-reward} all
operate on this object. Third, \emph{answer faithfulness}: the final answer is produced after
\texttt{</think>} and must remain consistent with, and derivable from, the graph; coupled with
the graph-utility reward, this makes the graph a load-bearing intermediate representation
rather than decorative scratch work.

\subsubsubsection{Stage 1: ORPO Cold Start}
We first align each backbone with Odds-Ratio Preference Optimization (ORPO)~\cite{orpo}, a
reference-free objective that combines a supervised negative-log-likelihood (NLL) term on the
preferred response $y_w$ with an odds-ratio penalty that suppresses the dispreferred response
$y_l$:
\begin{equation}
\mathcal{L}_{\text{ORPO}}
= \underbrace{-\log P_\theta(y_w \mid x)}_{\mathcal{L}_{\text{SFT}}}
\;+\; \lambda \,
\underbrace{-\log \sigma\!\left( \log \frac{\mathrm{odds}_\theta(y_w \mid x)}
{\mathrm{odds}_\theta(y_l \mid x)} \right)}_{\mathcal{L}_{\text{OR}}},
\qquad
\mathrm{odds}_\theta(y \mid x) = \frac{P_\theta(y \mid x)}{1 - P_\theta(y \mid x)},
\label{eq:orpo}
\end{equation}
where $\sigma$ is the logistic function and $\lambda$ weights the preference term. ORPO
collapses the usual supervised-fine-tuning-then-preference-optimization pipeline into a single
monolithic stage and requires no frozen reference model. We train for one epoch with a $5\%$
held-out split (seed $42$) for evaluation. This stage is the cold start in the sense of 
launching RL directly from a base model that almost never produces a well-formed 
trace yields a sparse, high-variance reward; establishing reliable format
adherence and a basic preference for good graph reasoning first makes the subsequent RL signal
dense and learnable~\cite{buehler_preflexor_2025}. The cold start thus plays a different role for each backbone: for the
already-reasoning Qwen3 models it primarily adapts existing reasoning ability to the
graph-native format, whereas for the standard Llama-3.2-3B-Instruct base it must induce the
reasoning behavior itself.

\subsubsubsection{Stage 2: Graph-GRPO}
Starting from the cold-start checkpoint, we apply Group Relative Policy Optimization
(GRPO)~\cite{shao_deepseekmath_2024}. For each prompt $q$ we sample a group of $G$ completions
$\{o_1,\dots,o_G\}$, score each with the composite reward $R(\cdot)$ of
\S\ref{subsec:gp-reward}, and form a critic-free, group-normalized advantage
\begin{equation}
\hat{A}_i = \frac{R(o_i) - \operatorname{mean}\big(\{R(o_j)\}_{j=1}^{G}\big)}
{\operatorname{std}\big(\{R(o_j)\}_{j=1}^{G}\big) + \varepsilon},
\label{eq:grpo-adv}
\end{equation}
used in the clipped policy-gradient objective with a KL penalty to a reference policy,
\begin{equation}
\mathcal{J}(\theta) = \mathbb{E}_{q,\{o_i\}}\!\left[
\frac{1}{G}\sum_{i=1}^{G} \min\!\Big( \rho_i \hat{A}_i,\;
\operatorname{clip}(\rho_i, 1-\epsilon, 1+\epsilon)\hat{A}_i \Big)
- \beta\, D_{\mathrm{KL}}\!\big(\pi_\theta \,\|\, \pi_{\text{ref}}\big)
\right],
\quad \rho_i = \frac{\pi_\theta(o_i \mid q)}{\pi_{\theta_{\text{old}}}(o_i \mid q)}.
\label{eq:grpo}
\end{equation}
Dispensing with a learned value network makes GRPO memory-efficient and well matched to our
setting, where the reward is a black-box composite of judge calls and graph analytics rather
than a differentiable signal. We sample $G=8$ completions per prompt, generate with a vLLM
backend~\cite{vllm} for throughput, and adapt only LoRA~\citep{lora} parameters to keep the
update lightweight and to mitigate catastrophic forgetting. Per-size budgets are in
Table~\ref{tab:gp-hparams}.

\subsubsection{Reward Definition}
\label{subsec:gp-reward}

Each completion $o$ receives a scalar reward that is a fixed convex combination of six
components, each normalized to $[0,1]$:
\begin{equation}
R(o) = \sum_{k} w_k\, r_k(o), \qquad
\mathbf{w} = (\,
\underbrace{0.30}_{\text{corr}},\,
\underbrace{0.15}_{\text{fmt}},\,
\underbrace{0.25}_{\text{util}},\,
\underbrace{0.10}_{\text{nx}},\,
\underbrace{0.10}_{\text{div}},\,
\underbrace{0.10}_{\text{struct}}\,), \qquad \textstyle\sum_k w_k = 1.
\label{eq:reward}
\end{equation}
Two components are model-graded by an external LLM judge
(\texttt{grok-4-1-fast-non-reasoning}) and four are computed programmatically from the parsed
graph $G=(V,E)$ with $n=|V|$ nodes and $m=|E|$ edges. In all reward figures the headline
``total reward'' is the trainer's per-step mean $\frac{1}{G}\sum_i R(o_i)$, equivalent to
Eq.~\eqref{eq:reward}.

\subsubsubsection{Semantic rewards (judge-graded)}
Let $J(a, a^\star)\in[0,1]$ denote the judge's continuous grade of a candidate answer $a$
against the gold answer $a^\star$. \emph{Correctness} grades the post-\texttt{</think>} answer
$a_o$ directly, $r_{\text{corr}} = J(a_o, a^\star)$. \emph{Graph utility} (the central
signal) first asks the judge to reconstruct an answer $\hat{a}_o = \textsc{Reconstruct}(G_o)$
using only the emitted \texttt{graph\_json} and no outside knowledge, then grades it,
$r_{\text{util}} = J(\hat{a}_o, a^\star)$. This is an information-bottleneck test: the model is
rewarded only when it has offloaded genuinely sufficient information into an explicit,
standalone graph.

\subsubsubsection{Format reward}
A graded check that the structured trace is present and the graph is parseable,
$r_{\text{fmt}} = \operatorname{clip}_{[0,1]}\!\big(\sum_{s\in\mathcal{S}} c_s\,
\mathbb{1}[s\text{ present}]\big)$, with section credits for \texttt{think} ($0.15$),
\texttt{brainstorm} ($0.10$), \texttt{graph} ($0.15$), a JSON-parseable \texttt{graph\_json}
($0.20$), \texttt{patterns} ($0.15$), \texttt{synthesis} ($0.15$), and a non-empty node set
($0.10$). Credit for \texttt{graph\_json} and everything after it is gated on the JSON parsing
successfully, so a malformed graph caps the score.

\subsubsubsection{NetworkX-validity reward}
With $E_{\text{inv}}$ the edges referencing nonexistent nodes and $\ell$ the number of
self-loops,
\begin{equation}
r_{\text{nx}} = \operatorname{clip}_{[0,1]}\!\Big(
0.3\,\mathbb{1}[n>0] + 0.3\,\mathbb{1}[E_{\text{val}}>0]
+ 0.2\big(1 - \tfrac{|E_{\text{inv}}|}{m}\big)
+ 0.1\,\mathbb{1}[\ell=0] + 0.1\,\mathbb{1}[\text{weakly connected}] \Big),
\end{equation}
$E_{\text{val}} = m - |E_{\text{inv}}|$, rewarding internally consistent, connected graphs.

\subsubsubsection{Diversity reward}
We embed the $m'$ textual graph elements (node ids and \texttt{source relation target}) with a Sentence-BERT model $\phi$ (\texttt{all-MiniLM-L6-v2})~\citep{sbert} and
measure the mean off-diagonal cosine similarity
\begin{equation}
\bar{s} = \frac{1}{m'(m'-1)} \sum_{k \neq l}
\frac{\phi(t_k)^\top \phi(t_l)}{\lVert \phi(t_k)\rVert\,\lVert \phi(t_l)\rVert},
\qquad
r_{\text{div}} = \operatorname{clip}_{[0,1]}\!\big(0.9\, d + b\big),\;\;
d = \operatorname{clip}_{[0,1]}\!\Big(1 - \tfrac{\bar{s} - 0.15}{0.35}\Big),
\end{equation}
with a richness bonus $b = \min(0.1, m'/100)$. This penalizes degenerate, collapsed graphs of
near-duplicate nodes that would otherwise reward-hack the validity and structure terms.

\subsubsubsection{Structure reward}
A topology score $r_{\text{struct}} = \operatorname{clip}_{[0,1]}(s_{\text{size}} +
s_{\text{dens}} + s_{\text{int}} + s_{\text{depth}} + s_{\text{conn}})$, with a size term
peaking for $5$--$20$ nodes; $s_{\text{dens}} = \min(0.2, 2\rho)$ for directed density
$\rho = \tfrac{m}{n(n-1)}$; an internal-node term $s_{\text{int}} = 0.3\,
\tfrac{n_{\text{int}}}{n}$ over nodes with both in- and out-edges; a depth term
$s_{\text{depth}} = \min(0.2, 0.2\tfrac{\min(L,6)}{6})$ for longest DAG path $L$; and a
weak-connectivity term ($0.1$). Internal nodes correspond to intermediate inferences and $L$
to reasoning-chain length, shaping the graph into a connected, hierarchical scaffold.

\subsubsubsection{Design rationale}
The bulk of the weight ($0.55$) is placed on the two semantic objectives (correctness and
graph utility) that capture the true task, while the four cheap programmatic terms ($0.45$)
act as dense shaping and anti-hacking rewards that keep every sampled trace valid, diverse,
and well-structured. Importantly, the components have heterogeneous effective maxima:
format and NetworkX-validity can reach $1.0$, whereas the diversity and structure terms are
soft-capped by design, a topically coherent graph cannot attain mean pairwise dissimilarity
($d\!<\!1$), and the internal-node and DAG-depth terms in $r_{\text{struct}}$ are mutually
competing. Their absolute levels are therefore not directly comparable, and the total reward
saturates below $1.0$ by construction.

\subsubsection{Training Dynamics and Results}
\label{subsec:gp-results}

Across all three sizes the cold start instills the reasoning format and a clear preference
ordering, and Graph-GRPO then improves the composite reward; the backbone type and size govern where each model starts and how much headroom remains.

\subsubsubsection{Graph-PRefLexOR-1.7B}
On a Qwen3-1.7B backbone (already a reasoning model with a native thinking mode) the ORPO
loss falls from $\sim\!2.11$ to $\sim\!1.38$, with the NLL
term essentially coincident with the total (the odds-ratio penalty is negligible under the
larger $5\times10^{-5}$ learning rate), while the preference accuracy rises from $0.95$ to
$1.0$ and the implicit reward margin grows steeply from $0.14$ to $\sim\!1.0$
(Fig.~\ref{fig:gp-orpo-panels}(a)), a far larger margin than the other sizes, reflecting the higher
learning rate rather than model scale. Under Graph-GRPO the composite reward improves modestly
from $\sim\!0.52$ to $\sim\!0.55$ over the $\sim\!970$-step run (Fig.~\ref{fig:gp-reward-panels}(a)),
with correctness rising $0.36\!\rightarrow\!0.42$, format and NetworkX-validity near their
ceilings, and graph utility the lowest component ($\sim\!0.22$).

\subsubsubsection{Graph-PRefLexOR-3B}
On the Llama-3.2-3B-Instruct backbone (a standard instruction model with no native reasoning, converted into a graph-native reasoner from scratch) the cold start converges cleanly within
one epoch: the
total ORPO loss and its NLL component fall from $\sim\!2.1$ to $\sim\!1.47$ and plateau after
$\sim\!300$ steps, separated by only $\sim\!0.02$--$0.03$ (a light odds-ratio regularizer),
while the preference accuracy reaches $\approx\!1.0$ and the reward margin grows from $0.03$ to
$0.16$, with the same accuracy on the held-out split (Fig.~\ref{fig:gp-orpo-panels}(b)). The modest
absolute margin reflects the deliberately large chosen/rejected gap built into the data
(\S\ref{subsec:gp-data}): the pairs are easy to separate, so this stage installs format and
preference adherence rather than final reasoning quality. Graph-GRPO then improves the
composite reward from $\sim\!0.46$ to $\sim\!0.55$ over the $\sim\!1970$-step (two-segment)
schedule (Fig.~\ref{fig:gp-reward-panels}(b)), representing the largest relative RL gain of the three, consistent
with the weakest cold start. The programmatic format and NetworkX-validity terms saturate
early; structure climbs $0.53\!\rightarrow\!0.64$ and diversity $0.67\!\rightarrow\!0.71$;
correctness improves only gradually ($0.34\!\rightarrow\!0.38$) and graph utility is the lowest
throughout ($0.17\!\rightarrow\!0.25$).

\subsubsubsection{Graph-PRefLexOR-8B}
On a Qwen3-8B backbone the ORPO loss falls from $\sim\!1.69$ to $\sim\!1.31$ in only
$\sim\!240$ steps, and the preference accuracy is $\approx\!1.0$ from the outset
(Fig.~\ref{fig:gp-orpo-panels}(c), the strong base separates preferences almost immediately, so the
cold start mainly instills the output format. Under Graph-GRPO the composite reward is high and
rises only modestly from $\sim\!0.60$ to $\sim\!0.63$, with a mid-run dip to $\sim\!0.60$
(Fig.~\ref{fig:gp-reward-panels}(c)). The component view explains why: the 8B leaves the cold start
already near the 3B's converged operating point; correctness sits at $\sim\!0.57$ and the
format, validity, diversity, and structure terms are high and stable, so RL consolidates and
lightly polishes rather than driving large gains; only graph utility drifts upward
($0.29\!\rightarrow\!0.32$) and stays the lowest component.

\subsubsection{Generation-length dynamics}
Beyond reward, Graph-GRPO reshapes the models' generation behavior
(Fig.~\ref{fig:gp-length-diag}(a)). The 1.7B and 8B models, which rarely exhaust their token budget
($\leq\!3\%$ truncated), lengthen their completed reasoning traces over training (to
$\sim\!2.6$k and $\sim\!2.4$k tokens respectively). The 3B exhibits the opposite and more
pronounced dynamic: it initially truncates up to $\sim\!20\%$ of completions against the
$8000$-token budget, and reinforcement learning drives this rate down to $\sim\!1\%$ while
simultaneously shortening the mean terminated length ($\sim\!2.7$k$\rightarrow\!2.3$k tokens).
Because a truncated trace cannot emit a well-formed closing answer, it is penalized by the
format and correctness terms and thus selected against; GRPO therefore teaches the policy to
complete its graph-native reasoning within budget, a regularization of length and termination
that the reward curve alone does not reveal.

\subsubsubsection{Optimization Diagnostics}
The within-group reward standard deviation (Fig.~\ref{fig:gp-length-diag}(b), top) sets the scale of the
GRPO advantage (Eq.~\ref{eq:grpo-adv}): the 3B begins with the largest dispersion
($\sim\!0.11$) and the 1.7B and 8B with roughly half that ($\sim\!0.05$--$0.06$), so the 3B
receives the strongest per-group learning signal (consistent with its larger reward gains) and
as its dispersion decays toward $\sim\!0.07$ the signal, and the gains, taper. Policy entropy
(bottom) decreases modestly for all models, indicating gradual sharpening; the 1.7B remains the
most exploratory and the 3B sharpens most. No run shows vanishing advantages (the fraction of
zero-variance groups stays at $0$), confirming the reward provides a usable gradient throughout.

\subsubsection{Cross-model Summary and Analysis}
Three trends are size-invariant. First, the cold start always achieves near-perfect preference
accuracy, confirming that ORPO reliably installs the structured trace before RL, aided by the
large chosen/rejected gap built into the data. Second, the headroom that Graph-GRPO exploits is
governed less by parameter count than by whether the backbone was already a reasoning model:
the Qwen3-1.7B and Qwen3-8B models, which begin with native reasoning ability, leave the cold
start close to saturation and are mostly consolidated by RL, whereas the Llama-3.2-3B-Instruct
model (a standard chat model converted into a reasoner from scratch) begins lowest, carries
the largest within-group reward dispersion (Fig.~\ref{fig:gp-length-diag}(b)), and shows the largest RL
gains. This is consistent with the 8B's high starting correctness and the 3B being the most
``moved'' by training despite its intermediate size. Third, and most importantly,
graph utility is the binding constraint at every scale ($0.21$--$0.32$): producing
knowledge graphs that are semantically sufficient to reconstruct the answer (not merely valid
and well-formed, which all models achieve) is the central open challenge of graph-native
reasoning, and the principal target for future scaling.

\begin{table}[t]
\centering
\caption{Training configuration for the Graph-PRefLexOR family. All sizes share the
reasoning-trace format and the six-component reward of \S\ref{subsec:gp-reward}.}
\label{tab:gp-hparams}
\begin{tabular}{lccc}
\hline
\textbf{Setting} & \textbf{1.7B} & \textbf{3B} & \textbf{8B} \\
\hline
Backbone & Qwen3-1.7B & Llama-3.2-3B-Instruct & Qwen3-8B \\
\hline
\multicolumn{4}{l}{\textit{Stage 1: ORPO cold start}} \\
\quad epochs / max length & 1 / 2048 & 1 / 2048 & 1 / 2048 \\
\quad learning rate & $5\times10^{-5}$ & $1\times10^{-5}$ & $1\times10^{-5}$ \\
\quad batch size & 1 & 1 & 2 \\
\hline
\multicolumn{4}{l}{\textit{Stage 2: Graph-GRPO}} \\
\quad learning rate & $5\times10^{-6}$ & $5\times10^{-6}$ & $5\times10^{-6}$ \\
\quad group size $G$ & 8 & 8 & 8 \\
\quad per-device batch / grad.\ accum. & 2 / 4 & 2 / 4 & 1 / 8 \\
\quad epochs & 1 & 1 & 3 \\
\quad max completion length & 8000 & 8000 & 3500 \\
\quad LoRA $(r,\alpha)$ & $(32, 64)$ & $(32, 64)$ & $(16, 32)$ \\
\hline
\end{tabular}
\end{table}

\begin{table}[t]
\centering
\caption{Models used in this study and corresponding Hugging Face identifiers.}
\label{tab:model_hf_ids}
\small
\begin{tabular}{lllll}
\hline
\textbf{Model} & \textbf{Scale} & \textbf{Base Model} & \textbf{Base type} & \textbf{Hugging Face ID} \\
\hline
Graph-PRefLexOR & 8B   & Qwen3-8B               & Reasoning & \texttt{lamm-mit/Graph-Preflexor-8b\_12292025} \\
Graph-PRefLexOR & 3B   & Llama-3.2-3B-Instruct  & Standard  & \texttt{lamm-mit/Graph-Preflexor-3b\_08012026} \\
Graph-PRefLexOR & 1.7B & Qwen3-1.7B             & Reasoning & \texttt{lamm-mit/Graph-Preflexor-1.7b\_08012026} \\
\hline
\end{tabular}
\end{table}

\subsection{Benchmark Question Generation}
\begin{figure}
    \centering
    \includegraphics[width=1\linewidth]{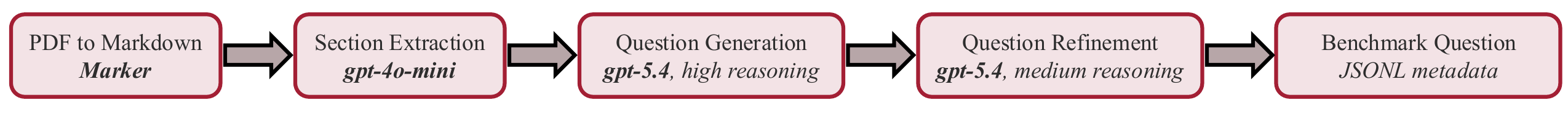}
    \caption{Workflow for constructing the open-ended scientific reasoning benchmark from research papers.}
    \label{fig:question_gen}
\end{figure}

The open-ended benchmark is constructed using a multi-stage pipeline for paper ingestion, section extraction, question generation, and question refinement (see Figure \ref{fig:question_gen}). We collect research papers spanning several domains relevant to scientific reasoning and materials design, including large language models, spider silk, polymer nanocomposites, epoxy networks, and collagen-based protein materials. Each PDF is converted to Markdown using the \texttt{Marker} library, which performs layout-aware text extraction without OCR or LLM-assisted parsing \cite{marker2025}.

Each Markdown file is then processed using OpenAI \emph{gpt-4o-mini} with a structured output schema to extract key paper-level fields, including the title, DOI, abstract, results, discussion, and conclusion \cite{openai_gpt4omini}. The Introduction, methods, and references sections are excluded because they typically contain background, procedural details, or citation metadata rather than the high-level mechanistic findings targeted by this benchmark. The extraction prompt includes robustness rules to account for variations in section naming, including \textit{Results and Discussion}, \textit{Concluding Remarks}, \textit{Summary}, and \textit{Discussion and Outlook}. The abstract, results, discussion, and conclusion sections are subsequently merged into a consolidated text block with explicit section headers. These records are compiled into a JSONL dataset containing the paper title, DOI, source text, and associated metadata. This consolidated representation serves as the input for benchmark question generation.

For each paper, OpenAI \emph{gpt-5.4} with high reasoning effort generates one self-contained, research-level evaluation question. The resulting benchmark contains 100 open-ended questions. Each question is assigned to one of five predefined reasoning categories: \texttt{causal\_multiscale\_reasoning}, \texttt{tradeoff\_and\_non\_monotonicity}, \texttt{hidden\_variable\_identification}, \texttt{model\_abstraction\_and\_breakdown}, or \texttt{cross\_domain\_mapping}. The generation prompt requires each question to define a clear scientific system, include relevant variables and mechanisms, and pose a non-trivial reasoning challenge involving a tradeoff, hidden variable, failure mode, or mechanistic breakdown. Additional constraints enforce a single-paragraph format, self-contained framing, a target length of 150--200 words, and a mechanistic final task beginning with either \emph{Explain why} or \emph{Then identify}.

Finally, each generated question is passed through a second \emph{gpt-5.4} refinement step to improve readability, grammar, precision, and benchmark suitability. This editing pass preserves the original system, variables, causal structure, question type, and intended reasoning challenge, while avoiding the introduction of new scientific claims or simplification of the task \cite{openai_gpt55}. The final output is a polished JSONL benchmark of 100 self-contained, open-ended scientific reasoning questions designed to evaluate mechanistic reasoning, causal inference, and hypothesis generation. Paper-level metadata, including DOI, title, and topic tags, are available at \url{https://huggingface.co/datasets/lamm-mit/graph-preflexor-grpo-benchmark}.

\subsection{Answer Backtracking and Hidden-State Analysis}
\label{sec:methods_backtracking}

We use semantic backtracking to measure which intermediate text is closest to a model's final answer. For each question $i$, let $a_i$ denote the final answer and let $\mathcal{C}_i = \{c_{i1}, c_{i2}, \ldots, c_{iK}\}$ denote the set of candidate reference texts. We embed the final answer and all candidate references using \texttt{BAAI/bge-base-en-v1.5} \cite{bge_embedding}. The BGE family provides general-purpose text embeddings for semantic retrieval and comparison~\cite{2023arXiv230907597X}, and sentence embeddings are commonly evaluated using cosine-based semantic similarity tasks~\cite{2019arXiv190810084R}. All embeddings are $\ell_2$-normalized before comparison.

For each candidate reference $c_{ij}$, we compute cosine similarity with the final answer:
\[
s_{ij} = \cos(e(a_i), e(c_{ij})) =
\frac{e(a_i)^\top e(c_{ij})}{\|e(a_i)\|_2 \|e(c_{ij})\|_2}.
\]
Because the embeddings are normalized, this is equivalent to the dot product between the two normalized vectors. The backtracking source is then assigned by top-1 nearest-reference selection:
\[
j^\ast_i = \arg\max_j s_{ij}.
\]
The source $c_{ij^\ast_i}$ is the candidate to which the final answer backtracks. We do not use a fixed absolute cosine-similarity threshold. Instead, each answer is assigned to the candidate source with the highest cosine similarity. This avoids choosing an arbitrary cutoff and makes the analysis fully reproducible from the embeddings and candidate list. If two candidates have exactly the same similarity up to numerical precision, we use a fixed deterministic ordering of candidates to break the tie.

For Qwen3-8B, we run the model with thinking enabled and split each output into a visible thinking trace and a final answer. We compare Qwen's final answer with six candidate references: Qwen's own thinking trace, Graph-PRefLexOR 8B's final answer, and Graph-PRefLexOR 8B's \texttt{<brainstorm>}, \texttt{<graph>}, \texttt{<patterns>}, and \texttt{<synthesis>} stages. We report both the full closest-source distribution and a binary split indicating whether Qwen's final answer backtracks to its own thinking trace or to another source.

For Graph-PRefLexOR 8B, we perform two related analyses. First, we compare Graph-PRefLexOR's final answer with its own reasoning stages and with Qwen3-8B outputs. The candidate set is:
\[
\{\texttt{<brainstorm>}, \texttt{<graph>}, \texttt{<patterns>}, \texttt{<synthesis>}, \text{Qwen thinking}, \text{Qwen answer}\}.
\]
This gives a cross-model backtracking distribution analogous to the Qwen analysis. We also report a binary split indicating whether Graph-PRefLexOR's final answer backtracks to its own reasoning stages or to Qwen3-8B outputs. Second, we perform an internal-only Graph-PRefLexOR analysis by comparing the final answer only with \texttt{<brainstorm>}, \texttt{<graph>}, \texttt{<patterns>}, and \texttt{<synthesis>}. This identifies which structured reasoning stage is closest to the final answer when cross-model references are removed.

We also analyze hidden states. For Qwen3-8B, we compute the mean hidden state over thinking tokens and the mean hidden state over final-answer tokens at each transformer layer. For layer $\ell$, with thinking-token positions $T_i$ and answer-token positions $A_i$, we compute
\[
h^{\mathrm{think}}_{i,\ell} = \frac{1}{|T_i|}\sum_{t \in T_i} h_{i,\ell,t},
\qquad
h^{\mathrm{ans}}_{i,\ell} = \frac{1}{|A_i|}\sum_{t \in A_i} h_{i,\ell,t}.
\]
The layer-wise thinking-answer divergence is then
\[
d_{i,\ell} = 1 - \cos(h^{\mathrm{think}}_{i,\ell}, h^{\mathrm{ans}}_{i,\ell}).
\]
For Graph-PRefLexOR 8B, we compute the analogous distance between hidden states from reasoning-stage tokens and hidden states from final-answer tokens. We also compute stage-specific distances by comparing each stage, \texttt{<brainstorm>}, \texttt{<graph>}, \texttt{<patterns>}, and \texttt{<synthesis>}, with the final-answer hidden state. All layer-wise plots report the mean across questions, with shaded bands showing one standard deviation.

Finally, we use two analyses to interpret the layer-wise peaks. First, we train a layer-wise linear probe to distinguish thinking-token hidden states from answer-token hidden states. This tests whether the distinction between reasoning and answering is linearly decodable at each layer. Second, we use a logit-lens-style approach ~\cite{2023arXiv230308112B} to project selected hidden states into vocabulary space and compare token preferences between reasoning and answer spans. These two analyses are observational. As a causal check, we use activation patching~\cite{2024arXiv240415255H}: hidden states from clean runs are patched into corrupted runs, and recovery is measured by final-answer similarity.

\subsection{Iterative Graph-native Ideation and Scaling Analysis}
\label{sec:methods-ideation}

We treat the graph-native reasoning model as a self-expanding ideation engine to study how additional test-time compute changes the structure of the generated idea space. Starting from a single seed topic, each iteration $t$ proceeds in three steps. First, the generator answers the current question and emits a structured reasoning trace, from which a parser extracts a local graph consisting of typed concept nodes and labeled relations. Second, this local graph is merged into a global directed graph using embedding-based de-duplication. Each new concept label is embedded with \texttt{google/embeddinggemma\_300m} \cite{vera_embeddinggemma_2025}, producing a 768-dimensional unit-normalized vector. The concept is merged with an existing canonical node if the cosine similarity is at least $0.85$; otherwise, it is added as a new node. Each node and edge is annotated with provenance metadata, including birth iteration $t$, reasoning depth, and originating response. Third, an expansion strategy inspects the accumulated graph and proposes a small set of follow-up questions, optionally anchored to selected nodes or node pairs, which define subsequent iterations. A best-first frontier orders the resulting work queue. The loop continues until the specified compute budget, defined by the number of model calls, total tokens, or iterations, is exhausted. Each generation is performed as an independent single-turn call; therefore, cross-turn memory is carried entirely by the accumulated graph rather than by the model context window. We run one experiment per expansion strategy using the common seed topic ''self-healing biopolymer composites''. The call, token, and iteration budgets are set to be effectively unbounded, and all analyses are performed on the first $2{,}000$ iterations of each run.

\subsubsubsection{Expansion Strategies}

The four runs differ only in the expansion policy that maps the accumulated graph to the next batch of questions. Each policy acts as an operator that allocates additional test-time compute to a different type of frontier in the evolving idea space. Let $G_t=(V_t,E_t)$ denote the accumulated graph at iteration $t$, let $\mathbf{x}_i\in\mathbb{R}^{768}$ be the unit-normalized embedding of node $i$, let $Q_t$ denote the set of nodes already used as anchors, and let $\deg(i)$ and $b(i)$ denote the degree and betweenness centrality of node $i$, respectively. The fan-out, or number of questions emitted per step, is denoted by $k$. The running concept centroid is defined as
\begin{equation}
\bar{\mathbf{x}}_t = \frac{1}{|V_t|}\sum_{i\in V_t}\mathbf{x}_i .
\end{equation}

The \emph{frontier} strategy is graph-analytic and balances outward expansion with consolidation of the graph core. It selects the lowest-degree unvisited nodes together with the single highest-betweenness hub,
\begin{equation}
\mathcal{L}_t=\big\{\,k\text{ nodes of smallest }\deg(i),\; i\in V_t\setminus Q_t\big\},
\qquad
h_t=\operatorname*{arg\,max}_{i\in V_t\setminus Q_t} b(i),
\end{equation}
and asks follow-up questions about unresolved mechanisms associated with each target. The \emph{novelty} strategy instead directs exploration toward the embedding periphery by selecting the $k$ nodes least aligned with the current centroid,
\begin{equation}
\mathcal{N}_t=\big\{\,k\text{ nodes of smallest }\mathbf{x}_i^{\top}\bar{\mathbf{x}}_t,\;
i\in V_t\setminus Q_t\big\}.
\end{equation}

The \emph{leap} strategy is deliberately divergent. For each of the $k/2$ most peripheral concepts $a$, it identifies the most semantically dissimilar partner anywhere in the graph,
\begin{equation}
p(a)=\operatorname*{arg\,min}_{j\in V_t,\,j\neq a}\;\mathbf{x}_a^{\top}\mathbf{x}_j,
\end{equation}
and generates a question that forces a concrete mechanistic connection between the pair. In parallel, for the $k$ most peripheral concepts, it imports a principle from an unrelated field, thereby encouraging cross-domain transfer and injecting concepts not yet present in the graph.

The \emph{converse} strategy operates at the level of language rather than graph structure. A separate questioner model, $\pi_{\mathrm q}$, reads the original seed question $q_0$ and the model's latest answer $a_t$ and proposes new questions,
\begin{equation}
\{q'_1,\dots,q'_k\}=\pi_{\mathrm q}(q_0,a_t),
\end{equation}
targeting implications, tensions, cross-domain analogies, or deeper mechanisms. Because this strategy reasons over the generated content rather than over existing graph nodes, it can introduce concepts absent from the accumulated graph and move beyond locally saturated regions. This comes at the cost of two model calls per iteration rather than one. The \emph{frontier}, \emph{novelty}, and \emph{leap} strategies generate node- or node-pair-anchored questions, whereas \emph{converse} generates unanchored follow-up questions. All strategies use the same generator, embedding model, de-duplication threshold, and compute budget, making the resulting runs directly comparable. Table~\ref{tab:strategies} summarizes the four policies.

\begin{table}[t]
\caption{Expansion strategies used for test-time graph expansion. Each strategy maps the accumulated graph to the next batch of questions, directing compute toward graph frontiers, embedding-peripheral concepts, distant recombinations, or language-level new directions. ''Calls'' denotes the number of model calls per iteration.}
\centering
\small
\begin{tabular}{@{}llll c@{}}
\toprule
Strategy & Targets selected per step & Driving signal & Search behavior & Calls \\
\midrule
\emph{frontier} & lowest-degree leaves $+$ top-betweenness hub & graph topology & widen frontier, consolidate core & 1 \\
\emph{novelty}  & nodes least aligned with centroid & embedding geometry & explore semantic periphery & 1 \\
\emph{leap}     & most-dissimilar pairs $+$ cross-domain transfer & embedding geometry & divergent recombination & 1 \\
\emph{converse} & new questions inferred from question and answer & questioner model & escape local saturation & 2 \\
\bottomrule
\end{tabular}
\label{tab:strategies}
\end{table}

\subsubsubsection{Scaling of Surprising-insight Yield} 

To quantify how insight yield scales with test-time compute, we reconstruct the accumulated graph at forty compute checkpoints, $t \in [0,2000]$. For each checkpoint, we filter the final graph to include only nodes and edges whose birth iteration is less than or equal to $t$, without re-running the model. All node labels are embedded once using \texttt{google/embeddinggemma\_300m}. At each checkpoint, we compute four metrics designed to be robust to graph size: the number of distinct concepts, which measures fluency; the total variance of node embeddings, which measures explored idea-space volume; the maximum embedding distance, $1-\cos$, between any node and the seed, which measures frontier reach; and the cumulative number of surprising recombinations, which serves as the primary insight-yield metric.

To define surprising recombinations, we estimate a global null distribution from the exact mean, $\mu$, and standard deviation, $\sigma$, of all pairwise cosine similarities in the final graph. A concept pair is defined as atypical when its combination score,
\begin{equation}
z_{\mathrm{comb}} = \frac{\cos(\mathbf{x}_i,\mathbf{x}_j)-\mu}{\sigma},
\end{equation}
is less than $-1$. We then count, cumulatively over $t$, all atypical concept pairs that the model has bridged through a shared intermediate concept. Specifically, we count pairs at graph distance two, connected through a common neighbor, but exclude directly linked pairs because direct edges are typically homophilic and therefore do not capture long-range recombination. Each bridged pair is assigned the first iteration at which it is realized, defined as the maximum birth iteration among its two endpoint nodes and the two connecting edges. For tractability on large graphs, distance-two enumeration is capped per hub. We avoid nearest-prior novelty because it is strongly confounded by graph size and mechanically decreases as the accumulated prior set grows. The resulting metrics are plotted against reasoning iteration for all four expansion strategies.

\subsubsubsection{Growth Dynamics} 

Whereas the scaling analysis reports cumulative quantities, the growth-dynamics analysis characterizes how a single graph develops over time by replaying the final graph in birth-iteration order. All measurements are based on embedding geometry or mesoscale community structure rather than raw graph distance, because shortest-path distances can shrink mechanically as the graph densifies.

We first replay nodes in order of arrival. Each newly added concept is classified as \emph{novel} if its embedding distance from the running concept centroid exceeds the median arrival distance, and as \emph{consolidating in-fill} otherwise. Concept arrivals are binned by iteration, and we compute the fraction of novel concepts in each bin. We also record the embedding distance of each new concept from the seed and report the per-bin mean and interquartile range as a measure of exploration radius.

We then replay edges in order across thirty checkpoints. At each checkpoint, we recompute the greedy-modularity community partition of the graph-so-far and report both the number of communities and the modularity $Q$. This provides a mesoscale view of how sub-fields form, proliferate, and interconnect after the graph becomes a single connected component. We define a recombination edge as any newly added edge whose endpoints were already connected by a prior path, and plot the embedding distance between its endpoints as a function of iteration, using a density plot with per-bin medians. An increasing median would indicate that later recombination edges bridge increasingly distant concepts. Finally, we track the degree trajectories of \emph{late bloomers}, defined as concepts that acquire the largest fraction of their final degree during the second half of compute. These concepts are selected by late-stage growth rather than by final degree alone. We report this dynamics analysis for the \emph{leap} run.

\section*{Statements and Declarations}

\subsection*{Funding}
This work was primarily supported by the U.S. Department of Energy, Office of Science, Office of Advanced Scientific Computing Research and Office of Basic Energy Sciences, Scientific Discovery through Advanced Computing (SciDAC) program under the FORUM-AI project. 

\subsection*{Competing Interests}
The authors declare that they have no competing financial or non-financial interests relevant to the content of this article.

\subsection*{Author Contributions}
M.J.B. and S.P. conceptualized the study and defined the project goals and investigation scope. M.J.B. performed model training and graph expansion analysis. S.P. designed the benchmark questions and carried out the PCA-based representation and semantic diversity analyses. S.S. conducted the semantic backtracking and layer-wise decomposition analyses. M.J.B. and T.G. provided project supervision and secured funding. All authors contributed to manuscript writing, review, and editing.

\subsection*{Data Availability}
The training dataset used in this study is available at \url{https://huggingface.co/datasets/lamm-mit/graph_reasoning_10K}. The benchmark data supporting the evaluation are available at \url{https://huggingface.co/datasets/lamm-mit/graph-preflexor-grpo-benchmark}. Additional model outputs, evaluation results, graph-expansion data, and analysis artifacts are available at \url{https://github.com/lamm-mit/graph-preflexor-grpo}, or will be made available from the corresponding author upon reasonable request.

\subsection*{Code Availability}
The full training and analysis code is available at \url{https://github.com/lamm-mit/graph-preflexor-grpo}.

\subsection*{Model Availability}
The trained Graph-PRefLexOR models are available at:
\url{https://huggingface.co/lamm-mit/Graph-Preflexor-8b_12292025},
\url{https://huggingface.co/lamm-mit/Graph-Preflexor-3b_08012026}, and
\url{https://huggingface.co/lamm-mit/Graph-Preflexor-1.7b_08012026}.

\subsection*{Use of Large Language Models}
Large language models were used for benchmark construction, teacher-response generation, rejected-response generation, question refinement, and independent evaluation, as described in the Methods section. All LLM-generated materials and outputs were reviewed, filtered, and analyzed by the authors.

\bibliographystyle{unsrt}
\bibliography{references}

\newpage


\section*{Supplementary material}

## S1 Representative Qwen3-8B Baseline Response

This section provides the full Qwen3-8B response to the representative benchmark question discussed in Section 2.2 of the main text. The response is included to show the linear, extended, and repetitive reasoning style of the baseline model compared with the structured phase-wise reasoning produced by Graph-PRefLexOR-8B.

**Qwen3-8B baseline response**

```
Okay, so I need to compare biological immune systems with multi-agent AI frameworks. Let me
start by recalling what I know about both.
First, the immune system: B-cells, T-cells, memory cells. They work together to detect
pathogens, respond, and adapt. B-cells produce antibodies, T-cells help in coordinating
the response, and memory cells remember past infections for faster response next time.
Communication happens through signals like cytokines, and there's a feedback loop where the
immune system learns from each encounter.
Then, multi-agent AI systems: planner, executor, critic. Planner generates solutions,
executor carries them out, critic evaluates and refines. They communicate through some
protocols, maybe messages or shared data. Adaptation comes from feedback loops where the
critic's evaluation informs the planner and executor.
So, the task is to map these components. Let me list the components:
Biological: B-cells, T-cells, Memory cells, communication (cytokines), adaptation (memory),
feedback loops (immune response cycles).
AI: Planner, Executor, Critic, communication protocols, adaptation (learning from evaluations),
feedback loops (evaluation to planning).
Correspondences: Maybe B-cells correspond to the planner? Or maybe the executor? Wait,
B-cells produce antibodies, which are like solutions. The critic could be similar to the
feedback mechanisms that evaluate the response. T-cells might be like the coordinators,
ensuring that the response is effective. Memory cells are like the stored knowledge or learned
experiences.
Communication pathways: In the immune system, cytokines are signals between cells. In AI,
maybe the communication is through shared data or messages between agents.
Adaptation mechanisms: The immune system adapts through memory cells and mutations. AI adapts
through learning algorithms, maybe reinforcement learning where the critic provides rewards or
penalties.
Feedback loops: In the immune system, after an infection, the system remembers and responds
faster. In AI, the critic evaluates the executor's performance, which feeds back to the
planner to adjust future plans.
Now, where does the analogy break down? Let me think. Learning in the immune system is
through experience, but it's not like machine learning. Memory cells are a form of memory,
but they don't learn in the same way. Also, the immune system has a more decentralized and
parallel processing, whereas AI systems might be more centralized.
In terms of long-term adaptation, the immune system can adapt over time through evolution of
the immune response, but AI systems might need explicit training data. Also, the immune system
can handle novel pathogens through generalization, while AI might need specific training for
new tasks.
So, the gap is in the learning mechanisms. The immune system's learning is more about
recognizing patterns and generating specific responses, whereas AI's learning is through
optimization algorithms. Also, memory formation in the immune system is biological and
physical, while in AI it's stored data.
Hypothesis: Next-gen multi-agent systems need a capability similar to the immune system's
memory and adaptability. Maybe a dynamic knowledge base that can evolve and generalize,
allowing agents to adapt to new problems without retraining from scratch. The mechanism could
involve a decentralized, evolving repository of solutions and patterns, with agents able to
query, modify, and integrate new information, similar to how memory cells store and recall
immune responses. This would allow for long-term robustness by continuously updating and
refining the knowledge base based on new experiences, akin to the immune system's memory and
adaptation.
```

## S2 Qwen3-8B Behavior Analysis on Layers 7–10

The layer-wise divergence analyses reveal a recurring increase in Qwen3-8B around layers 7-10. We interpret this region as an early transition zone in which the model begins to separate exploratory thinking representations from final-answer

representations. This does not imply that the final answer is stored in these layers. Rather, the result suggests that layers 7-10 mark a representational shift: the model has already entered an answer-generation mode, but the answer content is not yet fully consolidated. This distinction is important because probes and representation-distance analyses show what information is present or separable in hidden states, but do not by themselves prove that the model uses that information causally [1, 2].

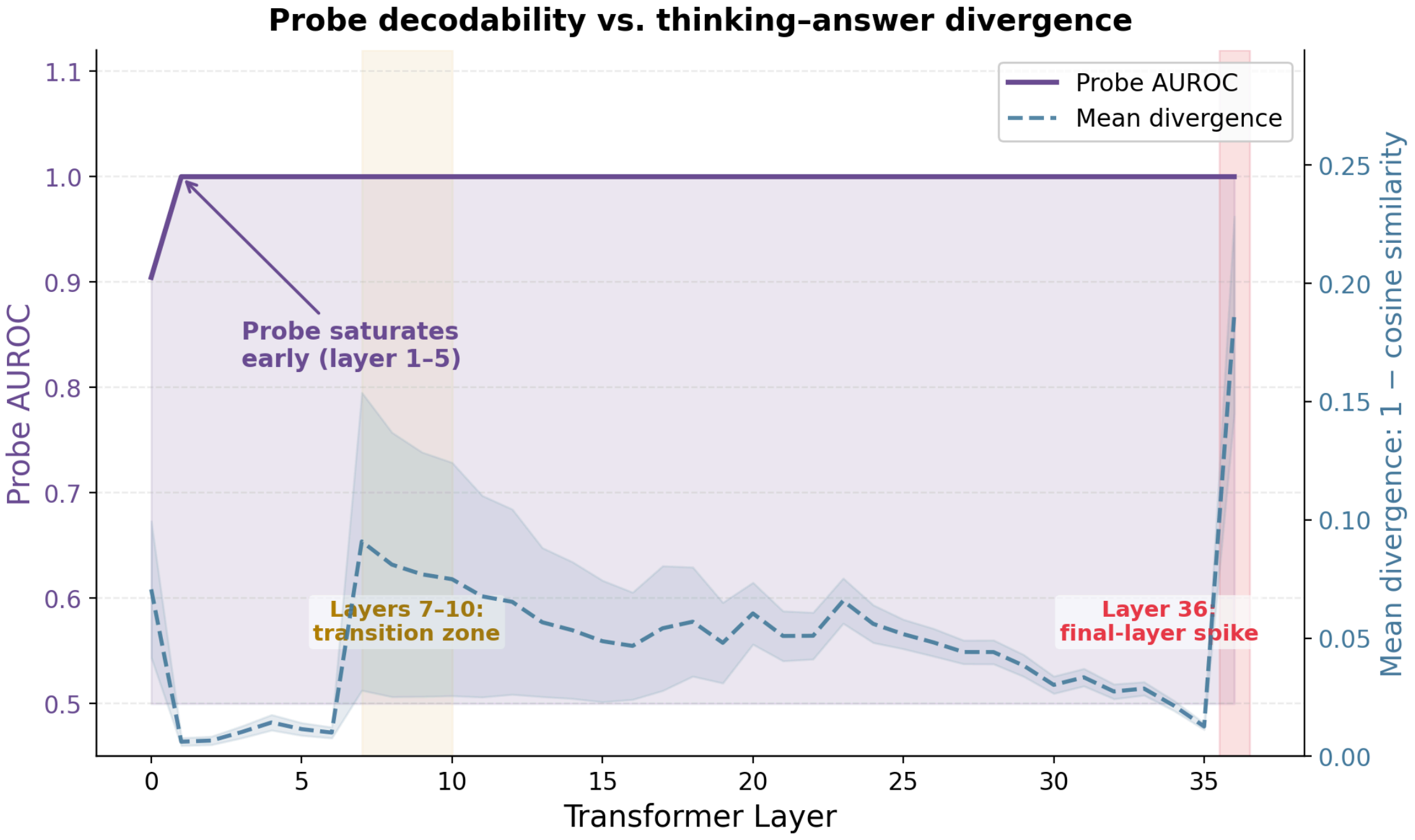


**Figure S1:** Probe decodability and thinking–answer divergence across Qwen3-8B layers. The linear probe reaches near-perfect AUROC before the layers 7–10 transition zone, showing that thinking-vs-answer identity is already linearly decodable early in the network. The hidden-state divergence curve nevertheless increases around layers 7–10 and spikes at the final layer, indicating stronger geometric separation between thinking and answer states. Thus, layers 7–10 are interpreted as a transition region rather than the location where final-answer content is stored.

Figure S1 separates two signals. The probe AUROC measures whether thinking-token and answer-token states are linearly separable. The divergence curve measures how far apart their average hidden-state representations are. The probe saturates early, while divergence rises later around layers 7-10. This suggests that Qwen3-8B can identify the generation mode before layers 7-10, but the two representation types become more geometrically separated in this region.

This is consistent with the linear-probe and activation-patching analyses. A standard linear probe can distinguish thinking-token states from answer-token states before layers 7-10, indicating that the model can identify the generation mode earlier in the network. However, the layer-wise cosine-distance profiles show that the hidden states become more strongly separated around layers 7-10. In this sense, layers 7-10 are not where the model first knows whether it is thinking or answering; they are where the two representation types become more geometrically separated. This agrees with prior work showing that transformer representations can change substantially across depth and that intermediate hidden states can expose evolving computational structure [3, 4, 5].

Activation patching further supports this distinction: patching layers 7-10 does not recover final-answer similarity as effectively as patching later layers. In our experiments, later layers, particularly around layer 30, recover final-answer similarity more strongly, suggesting that these later representations contain more directly recoverable answer content. The final layer exhibits the largest divergence, as expected, because it is closest to the output distribution and therefore most strongly shaped by final-answer wording.

Hence, these results clarify the role of the intermediate layers. Layers 7-10 appear to be diagnostic of the transition from visible reasoning to answer formation, whereas later layers carry more explicit answer-specific content. This distinction helps explain the stronger reasoning–answer divergence observed in Qwen3-8B: the final answer often moves away from the model's own visible thinking trace before being resolved in later layers. In contrast, Graph-PRefLexOR 8B shows a smaller reasoning–answer gap, and its final answers most often align with the `<synthesis>` stage. This

supports the central claim that graph-structured reasoning improves not only final-answer quality, but also the stability and interpretability of the pathway from intermediate reasoning to final response generation.



\end{document}